\documentclass[11pt,letterpaper]{article}
\usepackage[margin=1in]{geometry}
\usepackage[T1]{fontenc}
\usepackage[utf8]{inputenc}
\usepackage{times}
\usepackage[table]{xcolor}
\usepackage[most]{tcolorbox}
\usepackage{enumitem}
\usepackage{amssymb}
\usepackage{booktabs,graphicx,multirow,caption,wrapfig}
\usepackage{tabularx,rotating,colortbl}
\usepackage[authoryear,round]{natbib}
\setcitestyle{citesep={;},aysep={,},yysep={;}}

\usepackage{amsmath,amsfonts,bm}

\def\eqref#1{equation~\ref{#1}}

\def\1{\bm{1}}

\DeclareMathAlphabet{\mathsfit}{\encodingdefault}{\sfdefault}{m}{sl}
\SetMathAlphabet{\mathsfit}{bold}{\encodingdefault}{\sfdefault}{bx}{n}

\usepackage{url}
\usepackage[hidelinks]{hyperref}
\hypersetup{
  pdftitle={Decompose and Describe: Narrowing the Diagnosis-to-Report Gap in 3D CT Report Generation},
  pdfauthor={Zhisong Wang, Wanxing Chang, Weiwei Cao, Fenghe Tang,
             Bowen Shi, Ruifeng Yuan, Yong Xia, Jianpeng Zhang},
  pdfsubject={3D CT report generation},
  pdfkeywords={DeCAD, medical vision-language models, CT report generation}
}

\newcommand{\blue}[1]{#1}
\newcounter{finding}
\newtcolorbox{findingbox}{
  enhanced,
  colframe=blue!50!black,
  colback=blue!3,
  coltitle=black,
  boxrule=0.7pt,
  arc=5pt,
  left=5pt, right=5pt,
  top=5pt, bottom=5pt,
  fonttitle=\bfseries,
  step={finding},
  title={Finding \thefinding},
  attach title to upper,
  after title={:\ },
}

\title{\bfseries Decompose and Describe: Narrowing the Diagnosis-to-Report Gap in 3D CT Report Generation}
\author{%
  \normalsize
  Zhisong Wang\textsuperscript{1,2,3}\thanks{These authors contributed equally to this work.},
  Wanxing Chang\textsuperscript{1,2}\footnotemark[1],
  Weiwei Cao\textsuperscript{1,2},
  Fenghe Tang\textsuperscript{1,2,4},\\
  \normalsize
  Bowen Shi\textsuperscript{1,2},
  Ruifeng Yuan\textsuperscript{1,2},
  Yong Xia\textsuperscript{3}\thanks{Corresponding authors:
  \texttt{yxia@nwpu.edu.cn},
  \texttt{jianpeng.zhang0@gmail.com}.},
  Jianpeng Zhang\textsuperscript{1,2}\footnotemark[2]\\[0.6em]
\small \textsuperscript{1}DAMO Academy, Alibaba Group
\qquad \textsuperscript{2}Hupan Lab, Hangzhou, China\\
  \small \textsuperscript{3}Department of Radiology, Ningbo No. 2 Hospital,
  Ningbo, Zhejiang, China\\
  \small \textsuperscript{4}Jiangsu Key Laboratory of Multimodal Digital Twin Technology, Suzhou, Jiangsu, China
}
\date{}

\begin{document}

\maketitle

\begin{abstract}
Automated 3D CT report generation could support radiology workflows, yet medical vision--language models (MedVLMs) often omit diagnostic findings from generated reports. We ask whether the diagnostic ability of these models carries over to free-form reporting. Across five MedVLMs, diagnostic VQA achieves higher macro-F1 than reports generated from the same CT studies and evaluated on the same disease categories. We call this task-dependent discrepancy the \emph{diagnosis-to-report gap}. Conventional Report Adaptation improves reporting for most models but does not close the gap.
To investigate and narrow it, we introduce \emph{\textbf{DeCAD}} (\emph{Decompose CT Analysis and Describe}), which progressively decomposes CT report generation at diagnostic, regional textual, and regional visual levels. The model identifies positive findings before writing a report, organizes diagnosis and description into regional tasks, and restricts each task to visual tokens from the corresponding region. Diagnostic and regional textual decomposition improve report-level diagnostic performance across the five MedVLMs, whereas regional visual decomposition provides limited additional gains. In a standard VLM without medical VQA pretraining, the first two levels remain effective; regional visual decomposition also provides further gains with a region-level contrastively pretrained encoder. Full DeCAD achieves state-of-the-art performance on CT-RATE and Merlin.
Analyses show greater CT-token attention during diagnosis and higher recall of reference-positive findings in both the diagnostic output and final report than with Report Adaptation. These results show that explicitly organizing diagnosis improves diagnostic reporting and helps narrow the diagnosis-to-report gap.

\end{abstract}

\section{Introduction}
\label{sec:introduction}

Automated 3D CT report generation could support radiology workflows by reducing reporting workload~\citep{hamamci2024ct2rep,chen2025large}, but generating diagnostically accurate reports from volumetric scans remains challenging \citep{sloan2024automated,li2025automatic,chen2025large}. A report-generation model must not only identify abnormalities throughout a volume, but also select and organize relevant findings into coherent clinical descriptions. Prior studies have shown that fluent reports do not necessarily ensure factual completeness or diagnostic accuracy \citep{miura2021improving,chen2025dia}. 
Meanwhile, medical vision-language models (MedVLMs) have demonstrated increasingly strong image-interpretation capabilities \citep{li2023llava,xu2025lingshu,jiang2025hulu,hamamci2026generalist,liu2026radsight}, including on medical visual question answering (VQA) benchmarks \citep{lau2018dataset,zhang2023pmc,bai2024m3d}. Advances in medical image-text pretraining \citep{li2024anatomical,blankemeier2026merlin}, region-aware report generation \citep{tanida2023interactive,chen2025large}, and volumetric representations \citep{hamamci2026better} have further improved multimodal understanding. These developments raise a fundamental question: \emph{when a MedVLM can identify abnormalities under explicit diagnostic queries, does it reliably express the same information during free-form report generation?}

\begin{figure}
    \centering
    \includegraphics[width=1\linewidth]{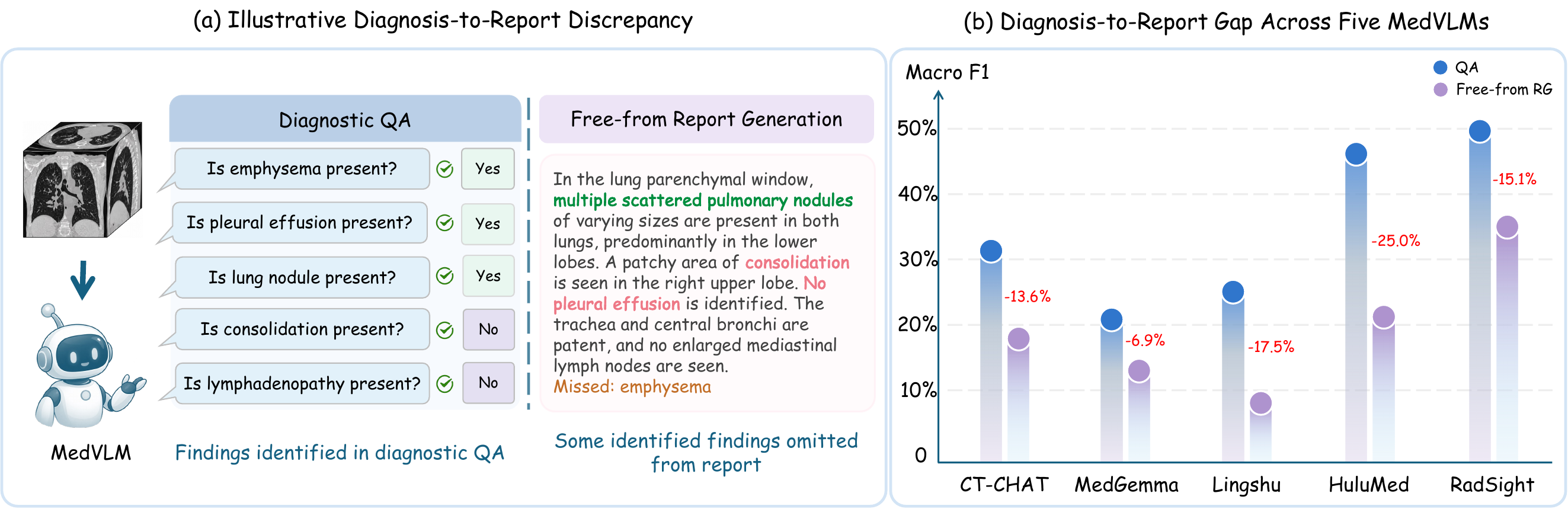}

    \caption{\textbf{Diagnostic ability and report expression in MedVLMs.}
    \textbf{(a)} Diagnostic QA responses and a free-form report generated from the same CT volume.
    \textbf{(b)} Diagnostic macro-F1 measured from QA answers and generated reports across five MedVLMs on CT-RATE.
    }
    \label{fig:teaser}
\end{figure}

We investigate this question by comparing diagnostic VQA and free-form report generation on the same CT studies using aligned disease categories and evaluation metrics. Diagnostic VQA consistently achieves higher macro-F1 across all five MedVLMs on CT-RATE (Fig.~\ref{fig:teaser}b), with a similar discrepancy observed on Merlin (Table~\ref{tab:medvlms_merlin}).
We term this task-dependent discrepancy the \emph{diagnosis-to-report gap}. The gap is particularly pronounced in recall, suggesting that free-form generation may fail to elicit and express findings that can be identified under explicit diagnostic prompts. Conventional fine-tuning on paired CT-report data, which we call \emph{Report Adaptation}, improves diagnostic reporting but does not eliminate this discrepancy (\textit{e.g.}, an 8.6-point macro-F1 gap for HuluMed-7B). 
These observations suggest that diagnostic capabilities elicited by explicit questions are not fully expressed in free-form reports, even after report-specific adaptation.

To investigate and narrow this gap, we introduce \textbf{\emph{DeCAD}} (\emph{\textbf{De}compose \textbf{C}T \textbf{A}nalysis and \textbf{D}escribe}), a progressive adaptation strategy that structures CT report generation through diagnostic, textual, and visual decomposition. 
At the diagnostic level, the model first identifies positive findings from the CT volume and then generates a report conditioned on both the image and this diagnostic assessment.
At the textual level, diagnostic assessment and report description are further organized into regional tasks, such as those for the lungs or heart, while retaining the full-volume CT as visual input.
At the visual level, regional visual decomposition allows the language model to access only the visual tokens associated with the requested region.
Across the evaluated MedVLMs, diagnostic and regional textual decomposition consistently improve reporting over direct image-to-report supervision, whereas regional visual decomposition provides limited additional benefit. These results suggest that direct reporting can underestimate the potential of existing MedVLMs: decomposition-guided adaptation can achieve state-of-the-art reporting performance without architectural modifications.

We next examine whether these benefits depend on the medical VQA pretraining of existing MedVLMs. When evaluated on a standard vision-language model without such pretraining, diagnostic and regional textual decomposition remain effective. This setting also allows us to revisit the gains from regional visual decomposition: with a region-level contrastively pretrained visual encoder, restricting visual conditioning to the requested region now provides further gains. Consequently, while diagnostic and textual decomposition remain effective without medical VQA pretraining, the efficacy of regional visual conditioning benefit from better alignment of regional visual representations.

To elucidate the mechanisms driving these reporting improvements, we probe the visual attention and finding expression within this standard VLM. Compared to conventional Report Adaptation, incorporating an explicit diagnostic assessment phase elicits stronger attention toward relevant CT visual tokens and significantly improves the recall of reference-positive findings—a benefit that cascades into the final generated report. These observations indicate that diagnostic decomposition systematically compels the model to ground its predictions in the visual input and articulate findings more comprehensively. Ultimately, this study demonstrates that optimizing the elicitation and structural organization of diagnostic information is as critical to automated multimodal reporting as advancing the core visual representations.

Our contributions are threefold:
\begin{itemize}[leftmargin=1.5em, labelsep=0.5em]
    \item We identify a \emph{diagnosis-to-report gap} across five MedVLMs through aligned evaluation of diagnostic VQA and free-form reporting on the same CT studies, and show that report-specific adaptation does not fully close it.
    \item We introduce \emph{DeCAD}, a progressive framework for analyzing and narrowing this gap through diagnostic, regional textual, and regional visual decomposition. Diagnostic and regional textual decomposition substantially improve reporting over direct image-to-report supervision without changing model architectures.
    \item We show that these benefits extend to a standard vision-language model without medical VQA pretraining. With a regionally aligned visual encoder, regional visual decomposition provides further gains, enabling full DeCAD to achieve state-of-the-art reporting performance.
\end{itemize}

\section{Do MedVLMs Report What They Can Diagnose?}
\label{sec:gap}

We compare diagnostic VQA and free-form reporting to examine whether findings identifiable under explicit diagnostic queries are reflected in generated reports. We then test whether report-specific adaptation is sufficient to close the observed gap.

\begin{findingbox}
    Existing MedVLMs exhibit a \emph{diagnosis-to-report gap}: they identify more labeled findings when answering explicit diagnostic questions than they express in free-form reports from the same CT studies. \emph{Report Adaptation} improves reporting for most models but does not close this gap.
\end{findingbox}

\subsection{Evaluation Protocol}
\label{sec:protocol}

\paragraph{Models and datasets.}
We evaluate five MedVLMs on CT-RATE~\citep{hamamci2026generalist} across 18 thoracic diagnostic categories: CT-CHAT~\citep{hamamci2026generalist}, MedGemma-1.5-4B~\citep{sellergren2026medgemma}, Lingshu-7B~\citep{xu2025lingshu}, HuluMed-7B~\citep{jiang2025hulu}, and RadSight-8B~\citep{liu2026radsight}. We additionally evaluate Merlin~\citep{blankemeier2026merlin}, HuluMed-7B, and RadSight-8B on the Merlin dataset across 30 abdominal diagnostic categories. We use official model checkpoints with their corresponding preprocessing and inference configurations.

\paragraph{Aligned evaluation.}
For each CT study, we compare diagnostic VQA and free-form reporting using the same disease categories and reference labels. Diagnostic VQA queries each category separately:
\textit{``Looking at these CT slices, is there \texttt{<disease>}? Answer yes or no.''}
For report generation, we use:
\textit{``Generate a comprehensive radiology report for this CT scan.''}
We extract disease predictions from generated reports using a publicly available RadBERT-based classifier for CT-RATE~\citep{yan2022radbert} and a ModernBERT-based classifier trained for Merlin~\citep{warner2025smarter}. Both tasks are evaluated using macro-averaged precision, recall, and F1, with macro-F1 as the primary metric.

\subsection{The Diagnosis-to-Report Gap in MedVLMs}
\label{sec:medvlm-gap}

We compare explicit diagnostic VQA with direct report generation. Under the aligned protocol, VQA achieves higher macro-F1 for all five models on CT-RATE, with gaps ranging from 6.9 to 25.0 points (Fig.~\ref{fig:teaser}b; Table~\ref{tab:medvlms_ctrate}). For HuluMed-7B, macro-F1 decreases from 46.3\% under VQA to 21.3\% under direct reporting, while recall decreases from 87.2\% to 15.4\%. Fig.~\ref{fig:teaser}a illustrates this discrepancy for an example CT volume.

The same pattern appears on abdominal CT. On Merlin, HuluMed-7B achieves a macro-F1 of 17.3\% under VQA versus 3.8\% under direct reporting, with recall of 78.1\% versus 4.0\% (Table~\ref{tab:medvlms_merlin}). Together, these results characterize the \emph{diagnosis-to-report gap}: although diagnostic information is accessible in visual representations and can be elicited by explicit queries, models express substantially less of this information in free-form reports.

\subsection{Is Report Adaptation Alone Enough?}
\label{sec:report-adaptation}

The diagnosis-to-report gap may partly reflect insufficient adaptation to report generation. We therefore fine-tune each model on paired CT-report data using direct image-to-report supervision, termed \emph{Report Adaptation}, while retaining its original architecture and visual input.

\textbf{Results.}
Report Adaptation improves diagnostic macro-F1 for most models (Tables~\ref{tab:medvlms_ctrate} and~\ref{tab:medvlms_merlin}). On CT-RATE, Lingshu-7B improves from 7.1\% to 21.5\% and HuluMed-7B from 21.3\% to 37.7\%, but all five models remain below their diagnostic VQA scores. On Merlin, HuluMed-7B improves from 3.8\% to 10.8\% and RadSight-8B from 19.8\% to 32.5\%, still trailing their VQA scores by 6.5\% and 7.9\%, respectively. Thus, direct report supervision narrows but does not close the diagnosis-to-report gap, motivating explicit diagnostic assessment before report composition.

\section{Can Task Decomposition Narrow the Diagnosis-to-Report Gap?}
\label{sec:decomposition}

\textit{Report Adaptation} leaves a residual diagnosis-to-report gap. We therefore examine whether decomposing report generation into intermediate diagnostic and regional tasks can further narrow this gap. \emph{DeCAD} progressively decomposes report generation at the diagnostic, textual, and visual levels. We first make diagnostic assessment explicit before report composition, then organize assessment and description by region, and finally examine whether restricting visual conditioning to the corresponding region provides additional benefits.

\begin{findingbox}
Across five MedVLMs, (a) explicit \emph{diagnostic decomposition} improves reporting over direct report supervision; (b) \emph{regional textual decomposition} provides further gains; and (c) \emph{regional visual decomposition} provides little incremental benefit with existing visual representations.
\end{findingbox}

\begin{figure}
    \centering

    \includegraphics[width=1\linewidth]{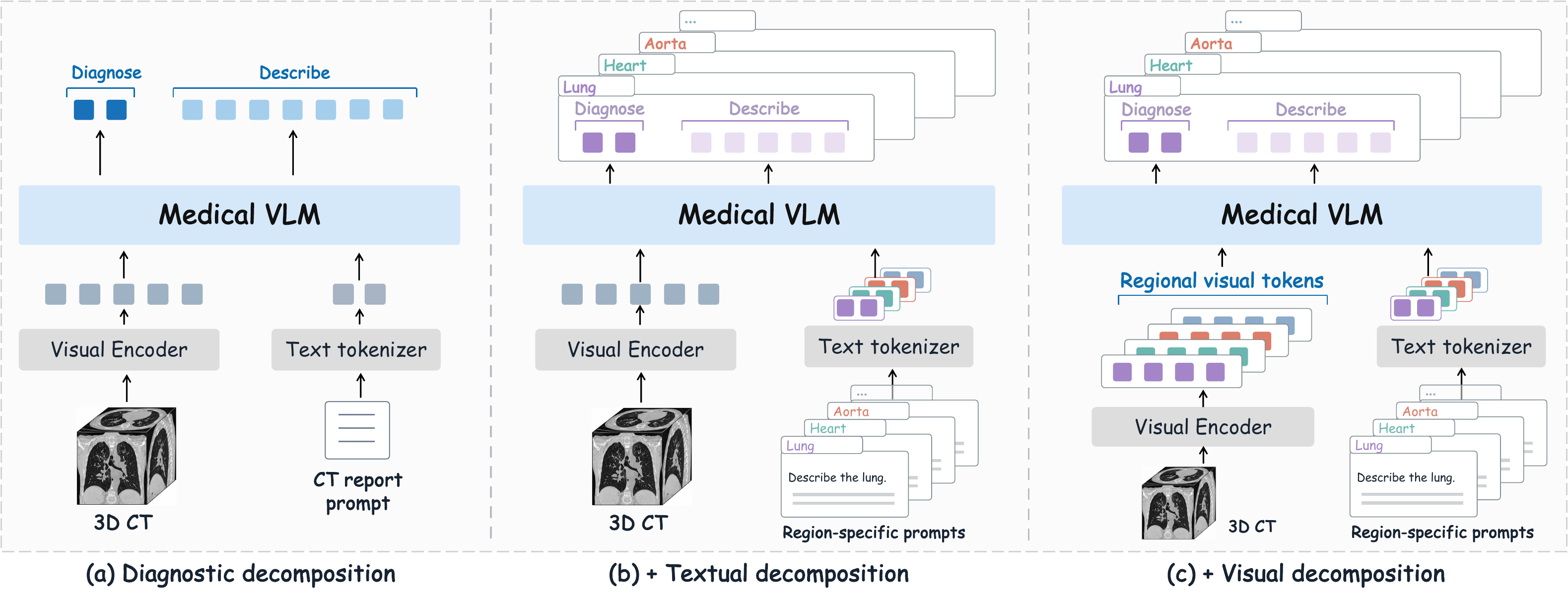}
    \caption{\textbf{DeCAD: progressive task decomposition for 3D CT report generation.}
    \textbf{(a)} Diagnostic decomposition generates an explicit diagnostic assessment before composing the report.
    \textbf{(b)} Regional textual decomposition organizes diagnostic assessment and report description by region, preserving the assessment-before-description sequence while retaining visual tokens from all target regions.
    \textbf{(c)} Regional visual decomposition further restricts each regional task to visual tokens associated with the requested region.}
    \label{fig:decompose}
\end{figure}

\definecolor{rankpurple}{HTML}{A486C8}
\definecolor{ranksecondblue}{HTML}{7FA8D4}

\begin{table*}[t]
\centering
\caption{Aligned diagnostic evaluation on CT-RATE. F1/P/R are macro-averaged diagnostic metrics evaluated against the same disease labels and reported as percentages. Within each model and metric, the best report-generation score is shaded purple and bold; the second-best distinct score is shaded blue and underlined. Ties share a rank. Diagnostic VQA is excluded from ranking.}
\label{tab:medvlms_ctrate}
\vspace{-3pt}

\begingroup
\scriptsize
\setlength{\tabcolsep}{3.2pt}
\renewcommand{\arraystretch}{1.15}

\def\rankfirst#1{\cellcolor{rankpurple!30}\textbf{#1}}
\def\ranksecond#1{\cellcolor{ranksecondblue!30}\underline{#1}}

\resizebox{\textwidth}{!}{%
\begin{tabular}{@{}l*{15}{c}@{}}
\toprule
& \multicolumn{3}{c}{\makebox[0pt][c]{CT-CHAT}}
& \multicolumn{3}{c}{\makebox[0pt][c]{MedGemma-1.5-4B}}
& \multicolumn{3}{c}{\makebox[0pt][c]{Lingshu-7B}}
& \multicolumn{3}{c}{\makebox[0pt][c]{HuluMed-7B}}
& \multicolumn{3}{c}{\makebox[0pt][c]{RadSight-8B}} \\
\cmidrule(lr){2-4}
\cmidrule(lr){5-7}
\cmidrule(lr){8-10}
\cmidrule(lr){11-13}
\cmidrule(l){14-16}
Setting
& F1 & P & R
& F1 & P & R
& F1 & P & R
& F1 & P & R
& F1 & P & R \\
\midrule

Diagnostic VQA
& 31.4 & 22.7 & 85.0
& 20.6 & 33.7 & 27.9
& 24.6 & 27.5 & 59.3
& 46.3 & 33.4 & 87.2
& 49.7 & 45.1 & 66.8 \\
\midrule

Direct
& 17.8 & 37.4 & 18.1
& 13.7 & 39.4 & 12.4
& 7.1 & 28.6 & 5.4
& 21.3 & 52.5 & 15.4
& 34.6 & 43.9 & 30.9 \\

Report Adapt.
& 17.7 & 39.7 & 15.1
& 19.8 & 46.0 & 19.5
& 21.5 & 56.2 & 17.8
& 37.7 & 53.8 & 33.2
& 37.2 & 43.8 & 33.7 \\

Diagnostic Decomp.
& 30.7 & \rankfirst{43.9} & \ranksecond{30.1}
& 32.3 & \rankfirst{50.7} & 28.9
& 30.8 & 47.2 & 27.8
& 45.0 & 57.4 & 42.9
& 42.2 & 45.3 & 40.6 \\

Diag.+Text Decomp.
& \rankfirst{36.6} & 40.2 & \rankfirst{42.0}
& \ranksecond{34.2} & \ranksecond{49.7} & \ranksecond{30.1}
& \ranksecond{33.4} & \ranksecond{56.6} & \ranksecond{29.5}
& \rankfirst{51.4} & \rankfirst{61.2} & \rankfirst{48.6}
& \ranksecond{45.8} & \ranksecond{46.1} & \ranksecond{49.2} \\

Diag.+Text+Visual
& \ranksecond{35.7} & \ranksecond{41.7} & \rankfirst{42.0}
& \rankfirst{34.3} & 49.4 & \rankfirst{32.3}
& \rankfirst{33.9} & \rankfirst{58.2} & \rankfirst{30.3}
& \ranksecond{49.9} & \ranksecond{60.4} & \ranksecond{48.1}
& \rankfirst{46.4} & \rankfirst{46.6} & \rankfirst{50.6} \\

\bottomrule
\end{tabular}%
}
\endgroup
\vspace{-5pt}
\end{table*}

\definecolor{rankpurple}{HTML}{A486C8}
\definecolor{ranksecondblue}{HTML}{7FA8D4}

\begin{table*}[t]
\centering
\caption{Aligned diagnostic evaluation on Merlin. F1/P/R are
macro-averaged diagnostic metrics evaluated against the same disease
labels and reported as percentages. Within each model and metric,
the best report-generation score is shaded purple and bold; the
second-best distinct score is shaded blue and underlined. Ties share
a rank. Diagnostic VQA is excluded from ranking.}
\label{tab:medvlms_merlin}
\vspace{-3pt}

\begingroup
\scriptsize
\setlength{\tabcolsep}{3.5pt}
\renewcommand{\arraystretch}{1.12}

\def\rankfirst#1{\cellcolor{rankpurple!30}\textbf{#1}}
\def\ranksecond#1{\cellcolor{ranksecondblue!30}\underline{#1}}

\begin{tabularx}{\textwidth}{@{}l*{9}{>{\centering\arraybackslash}X}@{}}
\toprule
& \multicolumn{3}{c}{Merlin}
& \multicolumn{3}{c}{HuluMed-7B}
& \multicolumn{3}{c}{RadSight-8B} \\
\cmidrule(lr){2-4}
\cmidrule(lr){5-7}
\cmidrule(lr){8-10}
Setting
& F1 & P & R
& F1 & P & R
& F1 & P & R \\
\midrule
Diagnostic VQA
& -- & -- & --
& 17.3 & 11.4 & 78.1
& 40.4 & 31.3 & 68.1 \\
\midrule
Direct
& 16.8 & \ranksecond{34.5} & 16.7
& 3.8 & 11.3 & 4.0
& 19.8 & 35.3 & 16.2 \\

Report Adapt.
& 25.5 & 24.5 & 29.1
& 10.8 & 21.1 & 9.5
& 32.5 & \rankfirst{38.1} & 30.8 \\

Diagnostic Decomp.
& 27.2 & 32.2 & 26.4
& 16.5 & 19.5 & 22.9
& \ranksecond{36.9} & 35.7 & \ranksecond{39.5} \\

Diag.+Text Decomp.
& \rankfirst{30.7} & \rankfirst{36.6} & \rankfirst{31.2}
& \rankfirst{23.9} & \ranksecond{29.0} & \ranksecond{26.2}
& \rankfirst{38.7} & 35.2 & \rankfirst{45.3} \\

Diag.+Text+Visual
& \ranksecond{29.0} & \ranksecond{34.5} & \ranksecond{29.5}
& \ranksecond{22.2} & \rankfirst{31.3} & \rankfirst{26.7}
& 33.4 & \ranksecond{37.3} & 35.1 \\
\bottomrule
\end{tabularx}
\endgroup
\vspace{-5pt}
\end{table*}

\subsection{Diagnostic Decomposition}
\label{sec:diagnosis-before-description}

Under direct report supervision, a model generates a complete report $r$ from a CT volume $x$:
\begin{equation}
    x \rightarrow r.
\end{equation}
Diagnostic assessment remains implicit within report generation. \emph{Diagnostic decomposition} makes it explicit by introducing an intermediate assessment $a$. During training, the assessment target is constructed from positive disease findings extracted from the reference report, and the model learns to generate the assessment before the report:
\begin{equation}
    x \rightarrow a \rightarrow r.
\end{equation}
The report remains conditioned on both the CT input and the preceding assessment. At inference, the model autoregressively generates both without externally supplied diagnostic labels. Relative to \emph{Report Adaptation}, this formulation retains the same architecture, visual input, CT-report training pairs, and report targets, while adding explicit diagnostic supervision before report composition.

\textbf{Results.} 
Diagnostic decomposition improves report generation over \emph{Report Adaptation} across all evaluated models. On CT-RATE, macro-F1 increases by 13.0, 12.5, 9.3, 7.3, and 5.0 points for CT-CHAT, MedGemma-1.5-4B, Lingshu-7B, HuluMed-7B, and RadSight-8B, respectively, with recall also increasing for all five models (Table~\ref{tab:medvlms_ctrate}). On Merlin, macro-F1 increases by 1.7, 5.7, and 4.4 points for Merlin, HuluMed-7B, and RadSight-8B, respectively. Recall increases by 13.4 and 8.7 points for the latter two (Table~\ref{tab:medvlms_merlin}). These results show that making diagnostic assessment explicit before report composition improves reporting across both chest and abdominal CT.

\subsection{Regional Textual Decomposition}
\label{sec:anatomical-decomposition}

We next examine whether organizing diagnostic assessment and report description by anatomical region provides further benefits. \emph{Regional textual decomposition} narrows each generation task to one region while preserving the order of diagnostic assessment followed by report description.

Let $\mathcal{R} = \{z_1, \ldots, z_K\}$ denote a predefined set of anatomical regions and $q_z$ the instruction for region $z$. Given the full CT volume $x$ and $q_z$, the model generates a regional diagnostic assessment $a_z$ followed by the corresponding report description $r_z$:
\begin{equation}
(x,q_z) \rightarrow a_z \rightarrow r_z,
\qquad z \in \mathcal{R}.
\end{equation}
The same model is trained on paired regional diagnostic assessments and report descriptions and invoked separately for each region at inference. Regional descriptions are then assembled into a complete report. All tasks retain the full-volume visual input, allowing us to isolate textual decomposition while holding the visual input fixed.

\textbf{Results.}
Compared with diagnostic decomposition, adding regional textual decomposition raises macro-F1 by 5.9, 1.9, 2.6, 6.4, and 3.6 points for CT-CHAT, MedGemma-1.5-4B, Lingshu-7B, HuluMed-7B, and RadSight-8B on CT-RATE, respectively. Recall increases by 1.2--11.9 points across all five models (Table~\ref{tab:medvlms_ctrate}). On Merlin, macro-F1 increases by 3.5, 7.4, and 1.8 points for Merlin, HuluMed-7B, and RadSight-8B, respectively, with corresponding recall gains of 4.8, 3.3, and 5.8 points (Table~\ref{tab:medvlms_merlin}). Thus, organizing diagnostic assessment and report description by region further improves reporting across both datasets without changing the visual input.

\subsection{Regional Visual Decomposition}
\label{sec:fine-grained}

Regional textual decomposition conditions each regional diagnostic assessment and report description on the full visual input. \emph{Regional visual decomposition} instead allows the language model to access only the visual tokens associated with the requested region. The language model and regional diagnostic assessment and report description targets remain unchanged.

\textbf{Results.}
Relative to regional textual decomposition, adding regional visual decomposition yields small macro-F1 changes on CT-RATE: $-0.9$, $+0.1$, $+0.5$, $-1.5$, and $+0.6$ points for CT-CHAT, MedGemma-1.5-4B, Lingshu-7B, HuluMed-7B, and RadSight-8B, respectively (Table~\ref{tab:medvlms_ctrate}). On Merlin, the corresponding changes are $-1.7$, $-1.7$, and $-5.3$ points for Merlin, HuluMed-7B, and RadSight-8B, respectively (Table~\ref{tab:medvlms_merlin}). Thus, regional visual decomposition provides little incremental benefit with the visual representations of these existing MedVLMs, motivating the representation-alignment analysis.

\begin{table}[t]
\centering
\caption{\textbf{Decomposition ablation without medical VQA pretraining.}
Diag., Text, and Visual indicate the three decomposition levels.
F1/P/R are macro-averaged percentages; bold marks the best result
in each column.}
\label{tab:ablation_training}
\vspace{-3pt}

\begingroup
\footnotesize
\setlength{\tabcolsep}{3.5pt}
\renewcommand{\arraystretch}{1.12}

\begin{tabular}{@{}lccc ccc ccc@{}}
\toprule
\multirow{2}{*}{Encoder pretraining}
& \multirow{2}{*}{Diag.}
& \multirow{2}{*}{Text}
& \multirow{2}{*}{Visual}
& \multicolumn{3}{c}{CT-RATE}
& \multicolumn{3}{c}{Merlin} \\
\cmidrule(lr){5-7}
\cmidrule(l){8-10}
& & & & F1 & P & R & F1 & P & R \\
\midrule

Region-level align.
& -- & -- & --
& 15.9 & 37.6 & 15.1
& 16.6 & 32.6 & 14.1 \\

Region-level align.
& $\checkmark$ & -- & --
& 39.0 & 45.7 & 39.0
& 23.2 & 26.2 & 27.7 \\

Region-level align.
& $\checkmark$ & $\checkmark$ & --
& 41.6 & 47.2 & 42.8
& 34.1 & 37.1 & 38.3 \\

Region-level align.
& $\checkmark$ & $\checkmark$ & $\checkmark$
& \textbf{52.1} & \textbf{55.6} & \textbf{54.2}
& \textbf{39.4} & \textbf{43.1} & \textbf{42.3} \\

\midrule

\textcolor{black}{Whole-image align.}
& $\checkmark$ & $\checkmark$ & --
& 46.9 & 51.5 & 45.8
& 33.6 & 30.7 & 39.0 \\

\textcolor{black}{Whole-image align.}
& $\checkmark$ & $\checkmark$ & $\checkmark$
& 46.0 & 54.0 & 44.1
& 30.9 & 29.4 & 38.9 \\

\bottomrule
\end{tabular}
\endgroup

\vspace{-3pt}
\end{table}

\definecolor{rankpurple}{HTML}{A486C8}
\definecolor{ranksecondblue}{HTML}{7FA8D4}

\begin{table*}[t]
\centering
\caption{\textbf{Report-generation comparison on CT-RATE.} All scores are
reported as percentages. F1/P/R are macro-averaged diagnostic metrics.
MET, RaTE, RG, and GR denote METEOR, RaTEScore, RadGraph-F1, and GREEN.
For each metric, the best distinct score is shaded purple and bold;
the second-best is shaded blue and underlined. Ties share a rank.
}
\label{tab:rpg_sota_ctrate}
\vspace{-3pt}

\begingroup
\footnotesize
\setlength{\tabcolsep}{3.6pt}
\renewcommand{\arraystretch}{1.12}

\def\rankfirst#1{\cellcolor{rankpurple!30}\textbf{#1}}
\def\ranksecond#1{\cellcolor{ranksecondblue!30}\underline{#1}}

\begin{tabularx}{\textwidth}{@{}l*{8}{>{\centering\arraybackslash}X}@{}}
\toprule
& \multicolumn{3}{c}{Diagnostic}
& \multicolumn{2}{c}{Text}
& \multicolumn{3}{c}{Clinical} \\
\cmidrule(lr){2-4}
\cmidrule(lr){5-6}
\cmidrule(l){7-9}
Method
& F1 & P & R
& MET & BLEU
& RaTE & RG & GR \\
\midrule

Lingshu-32B~\citep{xu2025lingshu}
& 9.3 & 19.7 & 8.4
& 10.4 & 1.0
& 50.4 & 5.7 & 10.8 \\

RadFM~\citep{wu2025towards}
& 11.6 & 19.5 & 11.2
& 3.7 & 0.1
& 34.8 & 0.8 & 1.3 \\

MedGemma-1.5~\citep{sellergren2026medgemma}
& 13.7 & 39.4 & 12.4
& 7.1 & 0.7
& 49.6 & 10.3 & 11.1 \\

Reg2RG~\citep{chen2025large}
& 17.1 & 39.6 & 12.5
& 24.6 & 23.5
& \rankfirst{66.9} & 25.2 & 37.9 \\

BTB3D~\citep{hamamci2026better}
& 17.3 & 21.0 & 15.9
& 20.4 & 15.8
& 59.0 & 15.1 & 22.2 \\

CT-CHAT~\citep{hamamci2026generalist}
& 17.8 & 37.4 & 18.1
& 23.6 & 23.6
& \rankfirst{66.9} & 22.2 & \rankfirst{42.2} \\

\textcolor{black}{Dia-LLaMA~\citep{chen2025dia}}
& 30.4 & 45.8 & 26.2
& 24.0 & \ranksecond{24.3}
& 66.2 & \ranksecond{25.7} & 41.7 \\

RadSight-8B~\citep{liu2026radsight}
& 34.6 & 43.9 & 30.9
& 23.0 & 21.2
& 65.7 & 21.6 & 38.6 \\

HuluMed-32B~\citep{jiang2025hulu}
& 38.0 & \ranksecond{48.0} & 37.0
& 22.5 & 17.9
& 64.1 & 17.0 & 39.8 \\

ViSD-Boost~\citep{cao2025boosting}
& 39.4 & 42.2 & 39.8
& \ranksecond{24.8} & 21.4
& 65.8 & 22.3 & 34.9 \\

AdaRAG-CT-8B~\citep{liang2026beyond}
& \ranksecond{41.8} & 45.1 & \ranksecond{46.7}
& 24.6 & 24.2
& 65.7 & \rankfirst{25.8} & 26.3 \\

\midrule

Ours (\emph{full DeCAD})
& \rankfirst{52.1} & \rankfirst{55.6} & \rankfirst{54.2}
& \rankfirst{26.3} & \rankfirst{27.1}
& \ranksecond{66.5} & 23.6
& \ranksecond{42.0} \\

\bottomrule
\end{tabularx}
\endgroup
\vspace{-3pt}
\end{table*}

\section{Does Task Decomposition Require Medical VQA Pretraining for 3D CT Report Generation?}
\label{sec:without-vqa}

Diagnostic and regional textual decomposition improve reporting in existing MedVLMs, but these gains may depend on prior medical VQA training. We therefore examine \emph{DeCAD} in a standard vision-language model without medical VQA pretraining. Its visual encoder is first trained with region-level CT-report contrastive learning to align regional visual representations with corresponding report descriptions. This setting allows us to test whether diagnostic and textual decomposition remain effective without medical VQA supervision and whether regional visual decomposition becomes beneficial with regionally aligned visual representations.

\begin{findingbox}
Diagnostic and regional textual decomposition remain effective without medical VQA pretraining. With a regionally aligned visual encoder, regional visual decomposition provides further gains. \emph{Full DeCAD} achieves state-of-the-art reporting performance.
\end{findingbox}

\subsection{A Standard Vision-Language Model and Training Setup}
\label{sec:simple-architecture}

Our standard VLM comprises a 3D ResNet-based visual encoder~\citep{hara2018can}, a projection layer, and a general-purpose Qwen3-4B language model~\citep{yang2025qwen3}. The visual encoder undergoes region-level CT-report contrastive pretraining to align regional visual features with corresponding report descriptions. It produces 32 visual tokens per target region, which the projection layer maps into the language model's embedding space.

We then freeze the visual encoder and jointly train the projection layer and language model using either \emph{Report Adaptation} or the decomposition formulations in Section~\ref{sec:decomposition}.
The architecture and pretrained visual encoder remain fixed across settings, allowing these formulations to be compared under the same visual representation. 
Further architecture and training details are provided in Appendix~\ref{sec:appendix_model}.

\subsection{Diagnostic and Textual Decomposition without Medical VQA Pretraining}

We compare \emph{Report Adaptation}, diagnostic decomposition, and diagnostic plus regional textual decomposition in the model without medical VQA pretraining (Section~\ref{sec:simple-architecture}). All settings share the same region-level contrastively pretrained visual encoder, language-model initialization, and CT-report training pairs. Each receives visual tokens from all target regions, allowing us to evaluate diagnostic and textual decomposition while holding the visual input fixed.

\textbf{Results.}
Diagnostic decomposition raises macro-F1 from 15.9\% to 39.0\% on CT-RATE and from 16.6\% to 23.2\% on Merlin (Table~\ref{tab:ablation_training}). Adding regional textual decomposition further raises these scores to 41.6\% and 34.1\%, gains of 2.6 and 10.9 points, respectively. Compared with Report Adaptation, the combined setting increases macro recall from 15.1\% to 42.8\% on CT-RATE and from 14.1\% to 38.3\% on Merlin. Macro precision also rises, from 37.6\% to 47.2\% and from 32.6\% to 37.1\%, respectively. Thus, diagnostic and regional textual decomposition improve the coverage of reference-positive findings without sacrificing precision, even without medical VQA pretraining.

\subsection{Regional Visual Decomposition with Region-Level Contrastive Pretraining}

Section~\ref{sec:fine-grained} showed that regional visual decomposition provides little incremental benefit with the visual representations of existing MedVLMs. We therefore revisit it in our standard VLM, whose visual encoder undergoes region-level image-text contrastive pretraining. Keeping the regional diagnostic assessment and report description tasks fixed, we restrict the language model to the 32 visual tokens associated with the requested region.

\textbf{Results.}
Adding regional visual decomposition to diagnostic and regional textual decomposition increases macro-F1 from 41.6\% to 52.1\% on CT-RATE and from 34.1\% to 39.4\% on Merlin, gains of 10.5 and 5.3 points, respectively (Table~\ref{tab:ablation_training}).
Table~\ref{tab:rpg_sota_ctrate} compares \emph{full DeCAD}, combining region-level contrastive pretraining with all three decomposition steps, against prior methods on CT-RATE. Full DeCAD achieves the highest macro-F1 and precision at 52.1\% and 55.6\%, respectively. Its macro-F1 exceeds that of the strongest compared prior method, AdaRAG-CT-8B~\citep{liang2026beyond}, by 10.3 points.
Full DeCAD also achieves the highest METEOR~\citep{banerjee2005meteor} and BLEU-4~\citep{papineni2002bleu} scores, surpassing the next-best results by 1.5 and 2.8 points, respectively. Clinical scores remain competitive, trailing the best RaTEScore~\citep{zhao2024ratescore}, GREEN~\citep{ostmeier2024green}, and RadGraph-F1~\citep{jain2021radgraph} by 0.4, 0.2, and 2.2 points, respectively. Overall, Full DeCAD achieves the strongest diagnostic macro-F1 while maintaining competitive text-generation quality and competitive clinical scores.

Together with the controlled ablations in Table~\ref{tab:ablation_training}, these results show that the three decomposition steps improve reporting with the region-level pretrained encoder in this model without medical VQA pretraining. In particular, under whole-image alignment, adding visual decomposition yields no improvement, reducing macro-F1 from 46.9\% to 46.0\% on CT-RATE and from 33.6\% to 30.9\% on Merlin. By contrast, under region-level alignment, visual decomposition further improves macro-F1, indicating that its benefits depend on region-level image–text alignment.

\begin{figure}
    \centering
    \includegraphics[width=\linewidth]{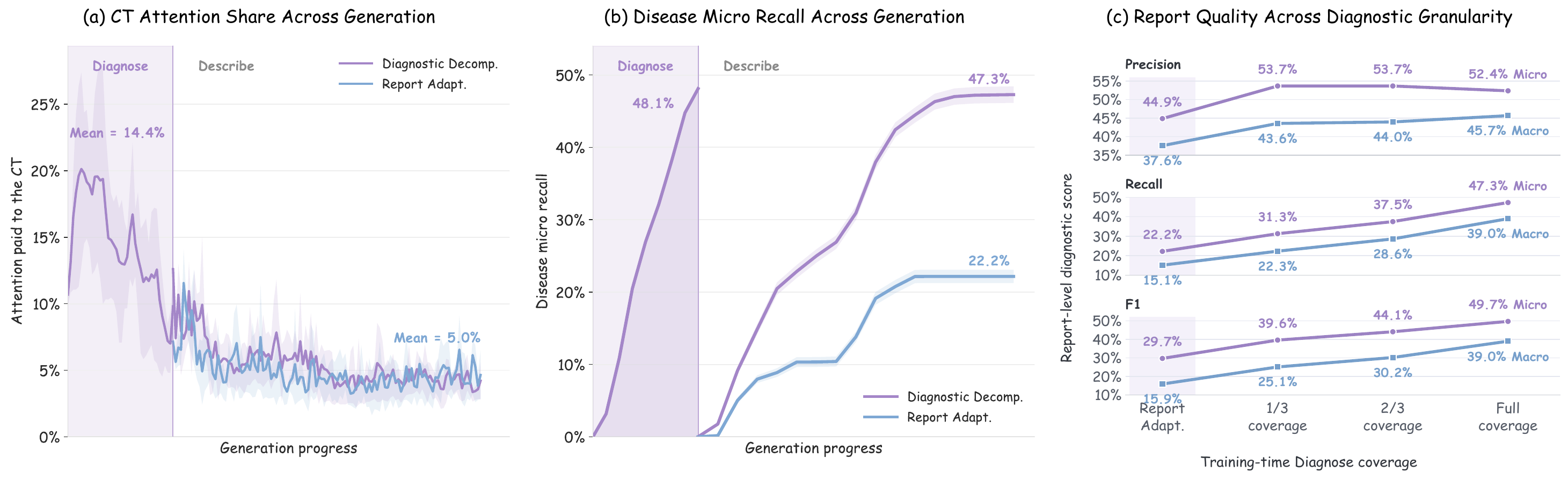}
    \caption{\textbf{Effects of Diagnostic Decomposition on report generation.}
    \textbf{(a)} Attention share assigned to CT visual tokens during \emph{Diagnose} and \emph{Describe}.
    \textbf{(b)} Micro recall of reference-positive findings across generated prefixes; the \emph{Describe} curve counts findings in the report alone.
    \textbf{(c)} Report-level precision, recall, and F1 under different levels of training-time \emph{Diagnose} coverage.}
    \label{fig:analysis}
\end{figure}

\subsection{How Does Diagnostic Decomposition Change Report Generation?}

\paragraph{Visual attention and finding expression.}
To examine how diagnostic decomposition may affect report generation, we analyze visual attention across generation stages. We teacher-force each model on its own generated output and measure the attention share assigned to CT visual tokens:
\begin{equation}
v_t=\frac{1}{H}\sum_{h=1}^{H}\sum_{j\in\mathcal{V}}\alpha_{t,j,h},
\end{equation}
where $H$ is the number of attention heads, $\mathcal{V}$ denotes visual-token positions, and $\alpha_{t,j,h}$ is the attention weight from position $t$ to position $j$ in head $h$ of the penultimate language-decoder layer. After resampling trajectories within each phase and averaging across cases, the mean visual attention share reaches $14.4\%$ during \emph{Diagnose}, compared with $5.0\%$ for Report Adaptation. During \emph{Describe}, it remains comparable to Report Adaptation (Fig.~\ref{fig:analysis}a). This pattern is consistent with greater reliance on CT visual tokens during explicit diagnostic assessment.

We next examine finding expression during the stage with increased visual attention. We track micro recall of reference-positive findings in generated prefixes, reading the \emph{Diagnose} list directly and using RadBERT for report text. The report curve resets at the onset of \emph{Describe} and counts only findings expressed in the report. Recall reaches $48.1\%$ during \emph{Diagnose}, compared with $22.2\%$ for Report Adaptation, and reaches $47.3\%$ in the subsequent report (Fig.~\ref{fig:analysis}b). Thus, the diagnostic phase exhibits both greater visual attention and more complete identification of reference-positive findings, while the resulting report retains substantially higher finding recall than \emph{Report Adaptation}.

\paragraph{Coverage of diagnostic supervision.}
We vary the proportion of training cases receiving explicit diagnostic assessment supervision, randomly selecting approximately one-third, two-thirds, or all cases. At inference, diagnostic assessments remain model-generated. Supervising one-third of the cases raises report-level micro precision from $44.9\%$ to $53.7\%$ relative to Report Adaptation (Fig.~\ref{fig:analysis}c). As supervision coverage increases, micro recall rises from $31.3\%$ to $37.5\%$ and $47.3\%$, while precision remains near $53\%$. These results are consistent with a role for explicit diagnostic supervision in improving finding precision, while broader supervision coverage is associated with higher recall of positive findings in reports.

\section{Related Work}

\paragraph{LLM-based MedVLMs.}
MedVLMs acquire medical image-understanding capabilities through domain-specific pretraining and instruction tuning. LLaVA-Med and Med-Flamingo explore biomedical instruction following and multimodal few-shot learning~\citep{li2023llava,moor2023med}, while RadFM, M3D, and CT-CHAT extend medical image understanding to volumetric inputs~\citep{wu2025towards,bai2024m3d,hamamci2026generalist}. Lingshu, HuluMed, and MedGemma 1.5 further expand task and modality coverage~\citep{xu2025lingshu,jiang2025hulu,sellergren2026medgemma}. Other efforts emphasize specific capabilities, such as segmentation and localization in VividMed~\citep{luo2025vividmed}, perception-oriented training in RadSight~\citep{liu2026radsight}, and diagnostic guidance in GSCo~\citep{he2026towards}. 
These advances have strengthened medical image understanding, yet it remains unclear how much of MedVLMs' diagnostic ability is reflected in free-form CT reports. 
We characterize this \emph{diagnosis-to-report gap} through aligned evaluation of diagnostic VQA and reporting on the same CT studies, and examine whether DeCAD improves the diagnostic content of reports generated by MedVLMs.

\paragraph{CT report generation.}
CT report generation has explored both visual representation learning and diagnostic guidance. CT2Rep generates reports from volumetric inputs using a visual encoder and autoregressive decoder~\citep{hamamci2024ct2rep}, while Argus investigates visual encoder pretraining, token compression, and scaling strategies~\citep{liu2025argus}. Reg2RG combines global context with anatomical region features and aligns regional representations with report descriptions~\citep{chen2025large}. AdaRAG-CT addresses incomplete pathology coverage by adaptively retrieving supplementary text and selectively incorporating it during generation~\citep{liang2026beyond}. For diagnostic guidance, Dia-LLaMA introduces disease prototype memory and disease-aware attention~\citep{chen2025dia}, while CT-AGRG predicts abnormalities before generating reports~\citep{di2025ct}. 
These studies improve CT reporting through visual representations, regional information, retrieval, and diagnostic guidance, but do not systematically examine the diagnostic gains from decomposing the generation process. DeCAD progressively decomposes diagnostic identification, regional textual organization, and visual conditioning to assess each level’s contribution to the model’s diagnostic performance and its final reports.

\section{Conclusion and Limitations}

We identified a \emph{diagnosis-to-report gap} in MedVLMs: diagnostic VQA consistently outperformed free-form reporting on aligned disease categories across five models on CT-RATE, with a similar discrepancy on Merlin. \emph{Report Adaptation} improved reporting for most models but did not close the gap. Through \emph{DeCAD}, we examined diagnostic, regional textual, and regional visual decomposition in sequence. Making diagnosis explicit before report composition improved reporting across the evaluated MedVLMs on CT-RATE, and organizing diagnosis and description by region provided further gains while retaining full-volume visual input. Regional visual decomposition added limited benefit in these models but provided further gains in a VLM with a visual encoder pretrained through region-level image--text contrastive learning. Without medical VQA pretraining, full DeCAD achieved the highest diagnostic macro-F1 among the compared report-generation methods on CT-RATE and Merlin, reaching 52.1\% and 39.4\%, respectively. These findings show that explicitly organizing diagnosis and description improves diagnostic reporting and helps narrow the diagnosis-to-report gap.

\paragraph{Limitations.}
Our diagnostic evaluation is restricted to the labeled disease categories in CT-RATE and Merlin and relies on RadBERT and ModernBERT for report labeling, respectively. 
Future work should extend evaluation to broader clinical settings and incorporate expert assessment.

\section*{Acknowledgments}
This work was supported by the Zhejiang Province Postdoctoral Research Excellence Funding Program (no. ZJ2024032). 

\newpage






\bibliographystyle{plainnat}
\bibliography{references}

\clearpage
\appendix
\section*{Appendix}

\section{Experimental Setup}
\label{sec:experimental_setup}

\paragraph{Datasets.}
We evaluate on two paired 3D CT-report datasets.
\textbf{CT-RATE} contains 25,692 non-contrast chest CT studies from 21,304 patients, paired with radiology reports and annotated with 18 thoracic abnormality categories.
\textbf{Merlin} contains 25,494 abdominal CT scans paired with radiology reports. We use its 30 common imaging findings as the diagnostic ontology for evaluation.
For both datasets, we follow the official data splits.

\paragraph{Existing MedVLMs.}
On CT-RATE, we evaluate HuluMed-7B, Lingshu-7B, CT-CHAT-8B, MedGemma-1.5-4B, and RadSight-8B. On the Merlin dataset, we evaluate the Merlin model, HuluMed-7B, and RadSight-8B. We follow each model's original CT preprocessing pipeline. HuluMed-7B uses its native 3D processor with 180 sagittal slices from a volume normalized to $[0,1]$. Lingshu-7B uses 40 axial slices after percentile-based intensity stretching, resizing to $448\times448$, and grayscale-to-RGB conversion. CT-CHAT-8B encodes CT volumes in Hounsfield units (HU) into visual latents offline. MedGemma-1.5-4B follows its HU-based input pipeline, while RadSight-8B follows its official preprocessing configuration using only the 3D visual pathway. For each model, the same input pipeline is used across diagnostic VQA and report-generation settings.

All fine-tuned variants of the existing MedVLMs are trained for two epochs with a learning rate of $2\times10^{-5}$, using a cosine learning-rate schedule with a warmup ratio of $0.03$, a per-GPU batch size of 1, and gradient accumulation over 8 steps. We use LoRA ($r=128$, $\alpha=256$) for Merlin, HuluMed-7B, Lingshu-7B, and RadSight-8B; LLaVA-style adaptation for CT-CHAT-8B; and full fine-tuning for MedGemma-1.5-4B.

\paragraph{Our standard VLM.}
Our standard VLM combines a 3D ResNet visual encoder with Qwen3-4B. We retain the original orientation of the preprocessed CT volumes, padding or cropping CT-RATE volumes to $112\times256\times352$ and Merlin volumes to $96\times256\times384$. Voxel intensities are clipped to $[-1000,1000]$ HU and linearly mapped to $[0,1]$.

Training proceeds in two stages. In Stage I, we pretrain the visual encoder for 10 epochs using either whole-image or region-level image--text contrastive learning, with a peak learning rate of $1\times10^{-4}$.
We then freeze the visual encoder and jointly train the visual projector and the full language model for five epochs with a learning rate of $2\times10^{-5}$. Stage-II training uses a cosine learning-rate schedule with a warmup ratio of $0.03$, a per-GPU batch size of 1, and gradient accumulation over 8 steps.

All training runs use four nodes with 32 NVIDIA H20 96GB GPUs in total.

\paragraph{Evaluation metrics.}
We evaluate generated reports along three dimensions. For diagnostic accuracy, we extract disease labels from the generated \texttt{<report>} body using a RadBERT-based report classifier on CT-RATE and a ModernBERT-based report classifier on Merlin, and compare them against the ground-truth labels. We report macro-averaged F1, precision, and recall, with macro-F1 as the primary metric. 
For text similarity, we report METEOR (MET) and BLEU-4 (BLEU). For clinical report quality, we report RaTEScore (RaTE), RadGraph-F1 (RG), and GREEN (GR), using the official GREEN-RadLlama2-7B evaluator~\citep{ostmeier2024green}.

\section{Data Construction and Disease Vocabularies}
\label{sec:appendix_data}

\subsection{Region-Level Report Decomposition and Disease Extraction}
\label{sec:disease_extraction}

\begin{figure}[t]
\centering
\includegraphics[width=1\linewidth]{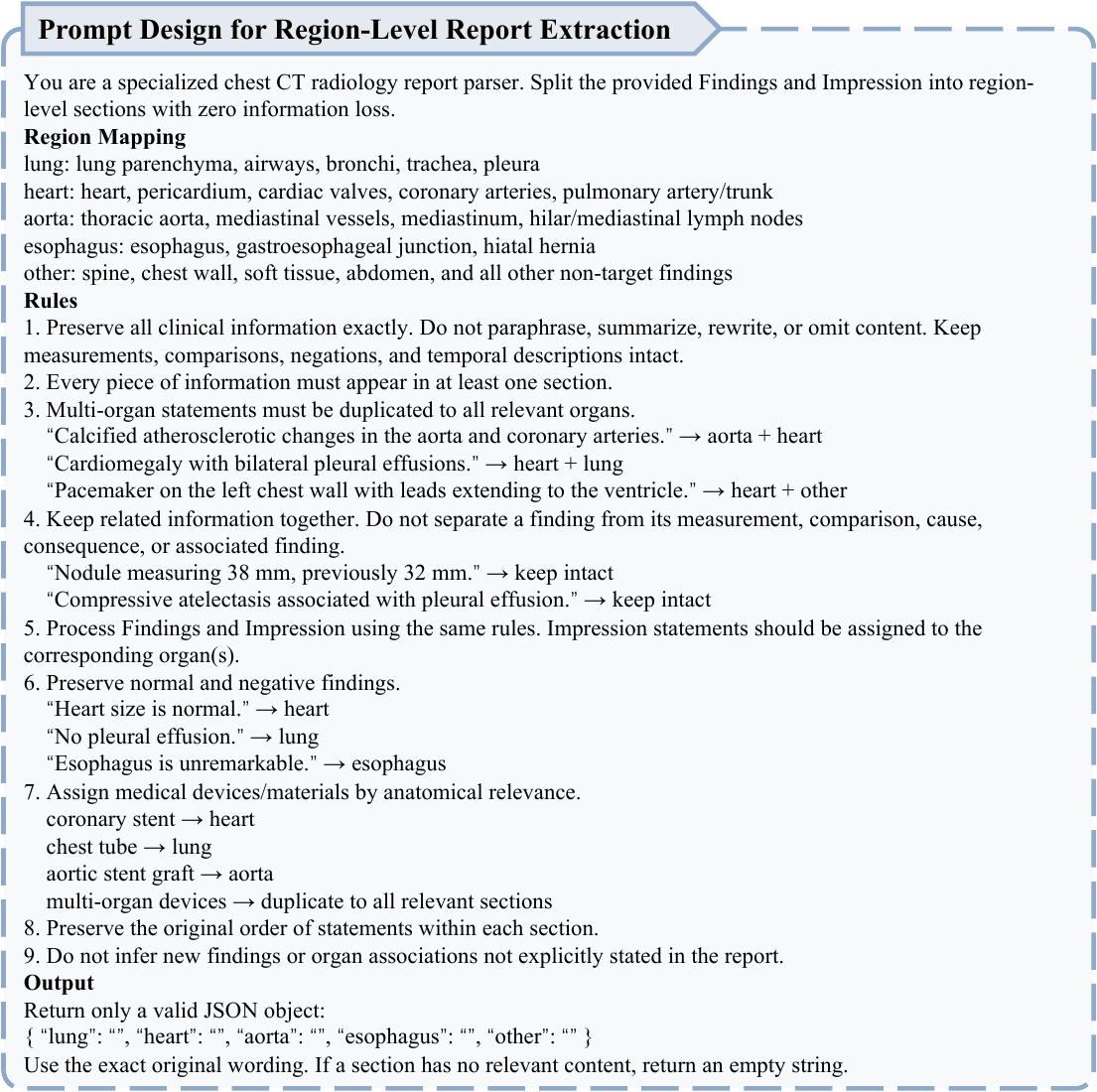}
\caption{\textbf{Prompt for region-level report decomposition.}
The prompt divides the original chest CT Findings and Impression into region-specific report sections for the lung, heart, aorta/mediastinum, and esophagus. It instructs the model to preserve the source clinical information, assign each statement to its relevant anatomical region, and repeat findings involving multiple regions when necessary. The resulting sections provide region-level textual supervision.}
\label{fig:organ_prompt}
\end{figure}

\begin{figure}[t]
\centering
\includegraphics[width=1\linewidth]{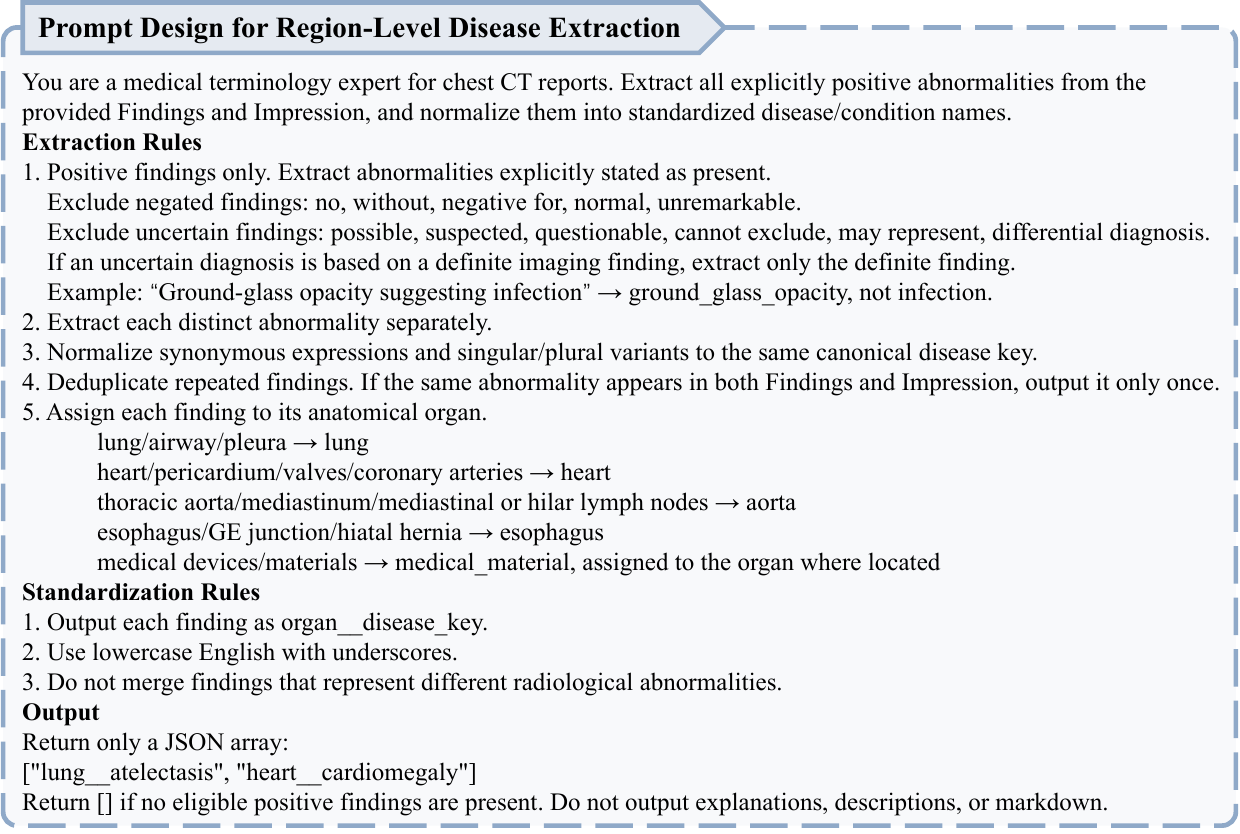}
\caption{\textbf{Prompt for region-level disease extraction.}
The prompt extracts explicitly positive abnormalities from chest CT Findings and Impression, assigns each finding to its relevant anatomical region, and normalizes synonymous expressions into standardized disease labels while excluding negated or uncertain findings and repeated mentions within each region.}
\label{fig:disease_prompt}
\end{figure}

To construct region-specific supervision, we adopt a two-stage processing pipeline for both CT-RATE and Merlin using Qwen-Max.
The expanded vocabularies described below provide detailed diagnostic supervision for model training, while diagnostic evaluation uses the 18 CT-RATE and 30 Merlin disease categories introduced in Sec.~\ref{sec:experimental_setup}.
We first decompose each radiology report into anatomical-region sections while retaining the clinical information relevant to each region. We then extract explicitly positive abnormalities from each regional section and normalize synonymous expressions into standardized disease labels. For clarity, Figs.~\ref{fig:organ_prompt} and~\ref{fig:disease_prompt} show the prompts instantiated for CT-RATE. Merlin follows the same procedure, with the region definitions and disease vocabulary adapted to its anatomical coverage.


\subsection{CT-RATE Disease Vocabulary}
For CT-RATE, we consider four anatomical regions and obtain \textbf{66 region-disease label assignments (64 unique disease labels)}: 42 for the lung, 12 for the heart, 9 for the aorta/mediastinum, and 3 for the esophagus.

\textbf{Lung (42):}
\textit{lung nodule, ground-glass opacity, atelectasis, emphysema, pulmonary fibrotic sequela, consolidation, pleural effusion, peribronchial thickening, bronchiectasis, calcified nodule, interlobular septal thickening, mosaic attenuation pattern, pneumonia, centrilobular nodule, crazy-paving pattern, pleural thickening, air cyst, vascular enlargement, ground-glass nodule, bulla, lung mass, metastatic nodule, pleural calcification, dependent density increase, cavitary lesion, honeycombing, halo sign, lung opacity, mucus plugging, pneumothorax, bronchiolitis, pulmonary edema, parenchymal calcification, fissure thickening, pleural nodule, intrapulmonary lymph node, tracheal diverticulum, tracheomegaly, lymphangitis carcinomatosa, pulmonary artery dilation, empyema}, and \textit{bronchial obstruction}.

\textbf{Heart (12):}
\textit{coronary artery wall calcification, cardiomegaly, pericardial effusion, pulmonary artery dilation, medical material, valve calcification, aortic valve calcification, atrial enlargement, ventricular dilation, mitral valve calcification, pericardial thickening}, and \textit{pulmonary hypertension}.

\textbf{Aorta/Mediastinum (9):}
\textit{arterial wall calcification, aortic dilation, calcified lymph node, lymphadenopathy, aortic aneurysm, medical material, aberrant subclavian artery, mediastinal mass}, and \textit{pneumomediastinum}.

\textbf{Esophagus (3):}
\textit{hiatal hernia, esophageal wall thickening}, and \textit{esophageal dilation}.

\subsection{Merlin Disease Vocabulary}
For Merlin, the same procedure is extended to its broader anatomical coverage. The resulting vocabulary contains \textbf{120 unique disease labels}, including \textbf{110 region-specific disease labels} across 15 anatomical regions and \textbf{10 additional study-level labels}. Unlike CT-RATE, each region-specific disease label in Merlin is assigned to a single anatomical region without duplication across regions.

\textbf{Adrenal Gland (5):}
\textit{adrenal nodule, adrenal adenoma, adrenal thickening, adrenal mass}, and \textit{surgically absent adrenal gland}.

\textbf{Aorta (5):}
\textit{abdominal aortic aneurysm, atherosclerosis, aortic calcification, aortic dissection}, and \textit{aortic occlusion}.

\textbf{Bladder (6):}
\textit{bladder wall thickening, bladder distension, bladder mass, bladder calculus, cystitis}, and \textit{surgically absent bladder}.

\textbf{Esophagus (5):}
\textit{hiatal hernia, esophageal dilation, esophageal wall thickening, esophageal varices}, and \textit{surgically absent esophagus}.

\textbf{Gallbladder (8):}
\textit{gallstones, surgically absent gallbladder, cholecystitis, gallbladder wall thickening, gallbladder distension, pericholecystic fluid, adenomyomatosis}, and \textit{biliary stent}.

\textbf{Heart (5):}
\textit{cardiomegaly, pericardial effusion, aortic valve calcification, mitral valve calcification}, and \textit{cardiac chamber enlargement}.

\textbf{Kidney (11):}
\textit{renal cyst, renal hypodensities, hydronephrosis, hydroureter, renal calculus, renal atrophy, renal mass, renal infarct, perinephric stranding, surgically absent kidney}, and \textit{delayed nephrogram}.

\textbf{Large Bowel (11):}
\textit{submucosal edema, diverticulosis, diverticulitis, colonic wall thickening, appendicitis, colonic mass, pneumatosis, rectal wall thickening, fecal loading, surgically absent appendix}, and \textit{surgically absent large bowel}.

\textbf{Liver (14):}
\textit{hepatic steatosis, hepatomegaly, hepatic cyst, hepatic metastasis, hepatic hemangioma, hepatic calcification, cirrhosis, biliary ductal dilation, intrahepatic biliary prominence, hepatic lesion, hepatic abscess, hepatic laceration, portal hypertension}, and \textit{hepatic atrophy}.

\textbf{Lumbar Vertebrae (5):}
\textit{fracture, compression fracture, degenerative changes, lytic lesion}, and \textit{sclerotic lesion}.

\textbf{Lung (8):}
\textit{atelectasis, pleural effusion, pulmonary nodule, consolidation, ground-glass opacity, bronchiectasis, pulmonary embolism}, and \textit{pneumothorax}.

\textbf{Pancreas (8):}
\textit{pancreatic atrophy, pancreatic calcification, pancreatic cyst, pancreatic mass, pancreatic duct dilation, pancreatitis, pancreatic fatty replacement}, and \textit{surgically absent pancreas}.

\textbf{Small Bowel (6):}
\textit{bowel obstruction, small bowel wall thickening, small bowel dilation, ileus, mesenteric stranding}, and \textit{surgically absent small bowel}.

\textbf{Spleen (6):}
\textit{splenomegaly, splenic lesion, splenic calcification, splenic infarct, surgically absent spleen}, and \textit{accessory spleen}.

\textbf{Stomach (7):}
\textit{gastric wall thickening, gastric distension, gastric mass, gastric bypass changes, gastric varices, gastric diverticulum}, and \textit{surgically absent stomach}.

\textbf{Additional Study-level Labels (10):}
\textit{ascites, anasarca, coronary calcification, free air, osteopenia, lymphadenopathy, metastatic disease, prostatomegaly, thrombosis}, and \textit{medical material}.

Together, the decomposed reports and standardized disease labels provide structured supervision for subsequent model training on both datasets.

\section{Model Architecture, and Training}
\label{sec:appendix_model}

Our standard VLM consists of a 3D visual encoder~\citep{hara2018can}, a projection layer, and a Qwen3 language model~\citep{yang2025qwen3}. Training is performed in two stages.
We first pretrain the visual encoder using either whole-image or region-level image--text contrastive learning. The region-level setting also uses auxiliary organ segmentation.
We then freeze the visual encoder and jointly optimize the projection layer and language model for report generation.

\subsection{Stage I: Visual Encoder Pretraining}
\label{sec:visual_encoder}

\paragraph{Region-level pretraining.}
The image branch employs a 3D ResNet-18 backbone to extract multi-scale volumetric features, together with an auxiliary segmentation branch for organ localization. The text branch uses BiomedVLP-CXR-BERT~\citep{boecking2022making}. Image and text representations are projected into a shared $d=256$ dimensional embedding space. The segmentation branch provides organ masks that localize each target region in the feature map. For organ $o$, we collect the spatial features within its corresponding mask as $X_o\in\mathbb{R}^{N_o\times d}$, where $N_o$ denotes the number of spatial features assigned to that organ.

A two-layer MLP predicts $M=32$ soft assignment maps, which aggregate these features into $M$ region-level visual tokens for organ $o$:
\begin{equation}
A_o=\mathrm{softmax}_{N_o}\bigl(f_o(X_o)\bigr),
\qquad
Z_o=A_o^\top X_o,
\end{equation}
where the softmax is applied over the $N_o$ spatial tokens independently for each of the $M$ slots, yielding $Z_o\in\mathbb{R}^{M\times d}$. A learned organ-specific query $\mathbf{u}_o$ further pools $Z_o$ into a single embedding for contrastive alignment with the corresponding organ description extracted from the report. Only organs fully contained within the input volume are included in the contrastive objective. The pretraining loss is
\begin{equation}
\mathcal{L}_{\mathrm{pre}}
=
\sum_o \mathcal{L}^{\mathrm{ITC}}_o
+
\mathcal{L}_{\mathrm{seg}},
\end{equation}
where $\mathcal{L}^{\mathrm{ITC}}_o$ denotes the region-level image-text contrastive loss for organ $o$~\citep{shui2025large}, and $\mathcal{L}_{\mathrm{seg}}$ is the Dice segmentation loss.

For region-level pretraining, we train a separate visual encoder for each dataset for 10 epochs, using a cosine learning-rate schedule with a peak learning rate of $10^{-4}$, weight decay of $0.05$, and a learnable contrastive temperature initialized to $0.07$.
The resulting encoders achieve mean AUCs of $0.789$ on CT-RATE and $0.824$ on Merlin; per-disease results are reported in Tables~\ref{tab:ctrate_disease_auc} and~\ref{tab:merlin_auc}. For downstream report generation, we retain the unpooled region-level visual tokens $Z_o$ for each organ, rather than the pooled contrastive embeddings.

\paragraph{Whole-image pretraining.}

\begingroup
For whole-image pretraining, we align each CT volume with its complete report using a symmetric image--text contrastive objective.
A learned global query aggregates spatial visual features into a whole-volume embedding. The report is encoded in non-overlapping text chunks, whose representations are averaged to obtain a report-level embedding.

Let $\mathbf{v}_i$ and $\mathbf{t}_i$ denote the normalized image and report embeddings for case $i$, respectively. For a global batch of $N$ paired samples, their similarity is $s_{ij}=\mathbf{v}_i^\top\mathbf{t}_j/\tau$, where $\tau$ is a learnable temperature initialized to $0.07$. Reports that are identical after lowercasing and whitespace normalization are
treated as additional positives. Let $\mathcal{P}_i$ be the set of positive indices for case $i$, including $i$ itself, and define $y_{ij}=1/|\mathcal{P}_i|$ if $j\in\mathcal{P}_i$ and $0$ otherwise.
The whole-image pretraining loss is

\begin{equation}
\mathcal{L}_{\mathrm{whole}}
=
-\frac{1}{2N}
\sum_{i=1}^{N}\sum_{j=1}^{N} y_{ij}
\left[
\log
\frac{\exp(s_{ij})}{\sum_{k=1}^{N}\exp(s_{ik})}
+
\log
\frac{\exp(s_{ji})}{\sum_{k=1}^{N}\exp(s_{ki})}
\right].
\end{equation}

The resulting encoders achieve mean zero-shot AUCs of $0.744$ on CT-RATE and $0.762$ on Merlin across the evaluated disease categories (Tables~\ref{tab:ctrate_disease_auc} and~\ref{tab:merlin_auc}). For downstream report generation, we retain the visual backbone and spatial projection and use the spatial tokens before global pooling.

\endgroup

\begin{table}[t]
\centering
\caption{\textbf{Zero-shot disease AUC on CT-RATE.} AUC across 16 disease categories for visual encoders pretrained with whole-image and region-level image--text contrastive learning, reported as percentages.}
\label{tab:ctrate_disease_auc}
\vspace{-3pt}

\begingroup
\small
\setlength{\tabcolsep}{8pt}
\renewcommand{\arraystretch}{1.05}

\begin{tabular}{@{}lcc@{}}
\toprule
Disease & \textcolor{black}{Whole-image} & Region-level \\
\midrule
Emphysema                          & 74.4 & 75.5 \\
Atelectasis                        & 65.6 & 65.3 \\
Lung nodule                        & 52.9 & 61.5 \\
Lung opacity                       & 25.6 & 71.5 \\
Pulmonary fibrotic sequela         & 60.7 & 64.8 \\
Pleural effusion                   & 89.5 & 81.5 \\
Mosaic attenuation pattern         & 81.0 & 85.3 \\
Peribronchial thickening           & 75.5 & 75.5 \\
Consolidation                      & 83.3 & 87.4 \\
Bronchiectasis                     & 71.1 & 76.6 \\
Interlobular septal thickening     & 85.3 & 72.1 \\
Cardiomegaly                       & 92.7 & 94.2 \\
Pericardial effusion               & 79.7 & 91.0 \\
Coronary artery wall calcification & 90.7 & 88.9 \\
Hiatal hernia                      & 71.4 & 80.4 \\
Arterial wall calcification        & 91.1 & 91.6 \\
\midrule
\textbf{Mean AUC}                  & \textbf{74.4} & \textbf{78.9} \\
\bottomrule
\end{tabular}

\endgroup
\end{table}

\begin{table*}[t]
\centering
\caption{\textbf{Zero-shot disease AUC on Merlin.} AUC across 30 disease categories for visual encoders pretrained with whole-image and region-level image--text contrastive learning, reported as percentages. Mean AUC is the unweighted average across disease categories.}
\label{tab:merlin_auc}
\vspace{-3pt}

\begingroup
\footnotesize
\setlength{\tabcolsep}{3pt}
\renewcommand{\arraystretch}{1.0}

\resizebox{\textwidth}{!}{%
\begin{tabular}{@{}lcc@{\hspace{10pt}}lcc@{}}
\toprule
Disease & \textcolor{black}{Whole-image} & Region-level
& Disease & \textcolor{black}{Whole-image} & Region-level \\
\midrule
Pleural effusion              & 95.6 & 97.1 & Renal cyst                  & 80.6 & 83.2 \\
Abdominal aortic aneurysm     & 77.9 & 96.8 & Appendicitis                & 58.1 & 82.9 \\
Bowel obstruction             & 88.1 & 95.5 & Submucosal edema            & 75.7 & 80.0 \\
Splenomegaly                  & 74.3 & 94.8 & Hepatomegaly                & 68.1 & 78.3 \\
Surgically absent gallbladder & 79.0 & 93.9 & Anasarca                    & 94.3 & 77.6 \\
Cardiomegaly                  & 86.0 & 93.8 & Ascites                     & 92.9 & 76.7 \\
Hydronephrosis                & 53.0 & 93.8 & Metastatic disease          & 61.0 & 75.6 \\
Atherosclerosis               & 84.0 & 89.8 & Free air                    & 80.2 & 71.9 \\
Aortic valve calcification    & 97.1 & 89.4 & Fracture                    & 77.3 & 71.5 \\
Hiatal hernia                 & 78.7 & 88.7 & Atelectasis                 & 82.3 & 70.8 \\
Pancreatic atrophy            & 79.1 & 87.7 & Renal hypodensities         & 76.0 & 70.1 \\
Biliary ductal dilation       & 74.3 & 86.3 & Osteopenia                  & 95.2 & 69.5 \\
Hepatic steatosis             & 45.8 & 86.1 & Thrombosis                  & 54.6 & 68.6 \\
Gallstones                    & 68.4 & 85.6 & Lymphadenopathy             & 53.8 & 67.9 \\
Coronary calcification        & 83.5 & 84.6 & Prostatomegaly              & 70.4 & 64.3 \\
\midrule
\textbf{Mean AUC} & \textbf{76.2} & \textbf{82.4} & & & \\
\bottomrule
\end{tabular}%
}

\endgroup
\vspace{-3pt}
\end{table*}

\subsection{Stage II: Report Generation Training}
\label{sec:report_training}

After visual pretraining, the encoder is frozen and a learnable projection layer maps the visual tokens into the embedding space of the language model. Under direct report supervision, the model is trained to generate the reference report directly from the CT input. Diagnostic decomposition instead trains the model autoregressively on a target sequence consisting of a diagnostic assessment of positive findings followed by the report.
For each target region, regional textual decomposition generates a diagnostic assessment before the corresponding report description.
All settings are optimized using autoregressive negative log-likelihood over their respective target sequences. At inference, Report Adaptation generates the report directly. Settings with diagnostic decomposition first generate diagnostic text and then generate the report conditioned on both the CT input and that text.

Regional visual decomposition restricts each regional task to its corresponding visual tokens. At inference, the model generates diagnostic assessments without external labels, and regional descriptions are concatenated to form the final report.
Intermediate assessments are excluded from report evaluation.

\subsection{Training and Inference Prompts}
\label{sec:prompt_details}

\begin{figure}[t]
\centering
\includegraphics[width=1\linewidth]{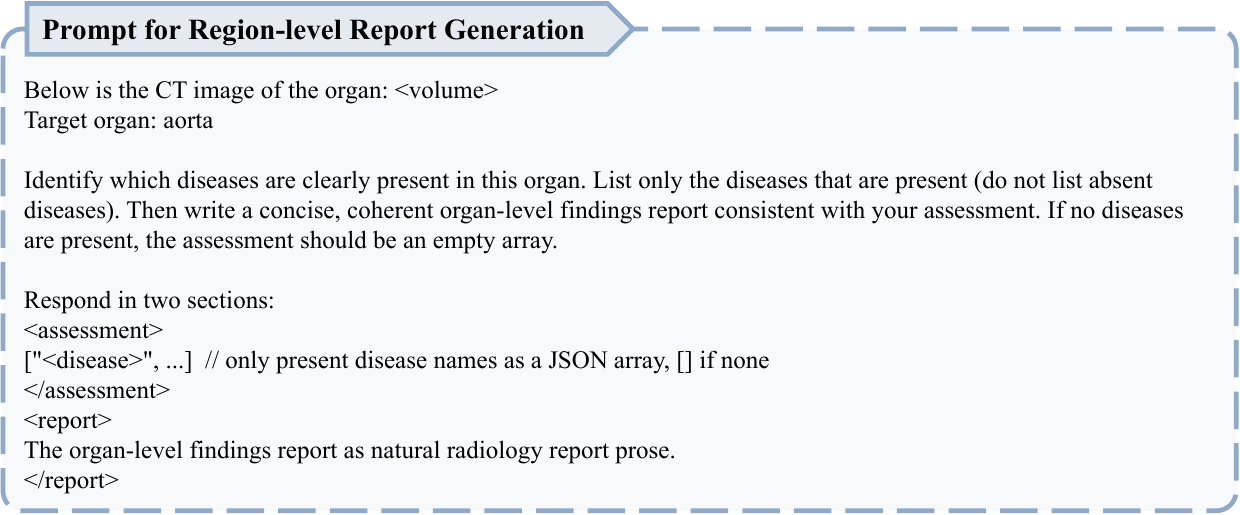}
\caption{Prompt for region-level report generation.}
\label{fig:train_region_prompt}
\end{figure}

\begin{figure}[t]
\centering
\includegraphics[width=1\linewidth]{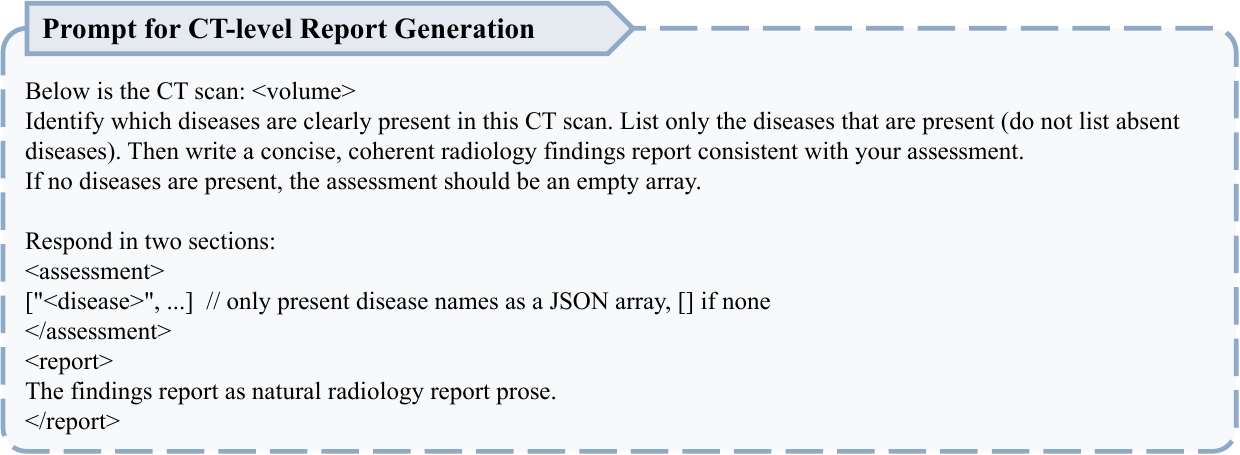}
\caption{Prompt for CT-level report generation.}
\label{fig:train_ct_prompt}
\end{figure}

Figures~\ref{fig:train_ct_prompt} and~\ref{fig:train_region_prompt} present the CT-level and region-level generation prompts, respectively. Each setting uses the same instruction during training and inference, requesting a complete report or a diagnostic assessment followed by a whole-volume or region-specific description. Regional visual decomposition uses the same prompts as regional textual decomposition; only the visual tokens supplied to each task differ. Assessments summarize positive findings; disease-specific yes/no questions are reserved for diagnostic VQA evaluation. Prompts for report decomposition and disease extraction are provided in Appendix~\ref{sec:disease_extraction}.

\section{Clinical-Efficacy Evaluation}
\label{sec:appendix_evaluation}

\subsection{Merlin Report Label Extractor}
\label{sec:merlin_modernbert}

\paragraph{Model.}
We use ModernBERT-base ($149$M parameters) as the report-level multi-label classifier. Each report is encoded by ModernBERT, and the \texttt{[CLS]} representation is passed through a dropout layer followed by a linear classification head that produces 30 independent logits. The model is optimized with binary cross-entropy loss.

\paragraph{Training.}
Reports are tokenized and truncated to a maximum length of $1024$ tokens. To address severe class imbalance, we apply a disease-specific positive-class weight,
$w_i = \mathrm{clip}(N_i^{\mathrm{neg}} / N_i^{\mathrm{pos}}, 1, 50)$,
in the binary cross-entropy loss, where $N_i^{\mathrm{pos}}$ and $N_i^{\mathrm{neg}}$ denote the numbers of positive and negative training examples for disease $i$, respectively. We also use sentence-shuffle augmentation, randomly permuting sentences within each report during training to reduce reliance on sentence order and encourage robustness to generated reports, which may present findings in a different order from reference reports. We train the model using AdamW with a learning rate of $2\times10^{-4}$, weight decay of $0.1$, a dropout rate of $0.1$, and a batch size of $32$. The learning rate follows a cosine annealing schedule with warm restarts. We use validation macro-F1 for early stopping and select the checkpoint with the highest score.

\paragraph{Results.}
On the held-out Merlin test set, the resulting extractor achieves a macro AUC of $98.8\%$ and a macro-F1 of $95.3\%$ across the 30 disease categories (Table~\ref{tab:merlin_extractor}), supporting its use as the report-label extractor for downstream CE evaluation.

\begin{table*}[t]
\centering
\caption{\textbf{Report-label extraction performance on Merlin.} Per-disease AUC and F1 on the held-out test set, reported as percentages.}
\label{tab:merlin_extractor}
\vspace{-3pt}

\begingroup
\small
\setlength{\tabcolsep}{5pt}
\renewcommand{\arraystretch}{1.05}

\begin{tabular}{@{}lrr@{\hspace{24pt}}lrr@{}}
\toprule
Disease & AUC & F1 & Disease & AUC & F1 \\
\midrule
Atelectasis                   & 100.0 & 100.0 & Metastatic disease          & 95.7 & 83.3 \\
Coronary calcification        & 100.0 & 100.0 & Osteopenia                  & 99.4 & 88.3 \\
Hepatic steatosis             & 100.0 & 100.0 & Fracture                    & 97.7 & 90.7 \\
Pleural effusion              & 100.0 & 99.9  & Abdominal aortic aneurysm   & 99.4 & 90.9 \\
Surgically absent gallbladder & 100.0 & 99.9  & Prostatomegaly              & 97.9 & 89.6 \\
Gallstones                    & 100.0 & 99.7  & Thrombosis                  & 86.1 & 89.6 \\
Renal cyst                    & 100.0 & 99.6  & Anasarca                    & 98.5 & 89.2 \\
Renal hypodensities           & 100.0 & 99.6  & Lymphadenopathy             & 99.0 & 85.1 \\
Splenomegaly                  & 100.0 & 99.6  & Bowel obstruction           & 99.8 & 91.8 \\
Cardiomegaly                  & 100.0 & 99.2  & Free air                    & 97.5 & 86.7 \\
Hepatomegaly                  & 100.0 & 99.1  & Submucosal edema            & 99.3 & 96.8 \\
Biliary ductal dilation       & 98.3  & 99.0  & Hydronephrosis              & 99.6 & 95.9 \\
Ascites                       & 99.9  & 98.9  & Hiatal hernia               & 99.9 & 97.5 \\
Pancreatic atrophy            & 99.9  & 98.7  & Appendicitis                & 99.6 & 94.1 \\
Atherosclerosis               & 98.1  & 98.9  & Aortic valve calcification  & 99.7 & 97.2 \\
\midrule
\multicolumn{4}{r}{\textbf{Macro average}} & \textbf{98.8} & \textbf{95.3} \\
\bottomrule
\end{tabular}

\endgroup
\end{table*}

\section{Additional Results and Analysis}
\label{sec:appendix_results}

\subsection{Micro-Averaged Clinical-Efficacy Results}
\label{sec:appendix_ce_results}
\definecolor{rankpurple}{HTML}{A486C8}
\definecolor{ranksecondblue}{HTML}{7FA8D4}

\begin{table}[h]
\centering
\caption{Micro-averaged diagnostic performance on CT-RATE (\%),
computed using RadBERT. Within each model and metric, the best
report-generation score is shaded purple and bold; the second-best
distinct score is shaded blue and underlined. Diagnostic VQA is
excluded from ranking.}
\label{tab:ct_rate_micro}

\begingroup
\footnotesize
\setlength{\tabcolsep}{3pt}
\renewcommand{\arraystretch}{1.10}
\def\rankfirst#1{\cellcolor{rankpurple!30}\textbf{#1}}
\def\ranksecond#1{\cellcolor{ranksecondblue!30}\underline{#1}}

\begin{tabularx}{\linewidth}{@{}l*{9}{>{\centering\arraybackslash}X}@{}}
\toprule
& \multicolumn{3}{c}{CT-CHAT}
& \multicolumn{3}{c}{MedGemma-1.5-4B}
& \multicolumn{3}{c}{Lingshu-7B} \\
\cmidrule(lr){2-4}
\cmidrule(lr){5-7}
\cmidrule(l){8-10}
Setting
& F1 & P & R
& F1 & P & R
& F1 & P & R \\
\midrule
Diagnostic VQA
& 31.9 & 19.8 & 82.2
& 30.5 & 31.1 & 29.9
& 30.1 & 20.7 & 54.5 \\
\midrule
Direct
& 30.1 & 38.1 & 24.9
& 15.9 & 35.3 & 10.2
& 11.4 & 39.5 & 6.6 \\
Report Adapt.
& 28.1 & \rankfirst{44.3} & 20.6
& 38.3 & 51.3 & 30.5
& 31.8 & 55.7 & 22.3 \\
Diagnostic Decomp.
& 39.8 & 42.2 & 37.7
& \ranksecond{44.9} & \ranksecond{55.9} & \ranksecond{37.5}
& 41.3 & 55.1 & 33.0 \\
Diag.+Text Decomp.
& \rankfirst{45.7} & \ranksecond{43.2} & \ranksecond{48.6}
& 44.4 & \rankfirst{56.1} & 36.8
& \ranksecond{44.3} & \ranksecond{61.5} & \ranksecond{34.6} \\
Diag.+Text+Visual
& \ranksecond{44.5} & 40.3 & \rankfirst{49.6}
& \rankfirst{47.1} & 55.5 & \rankfirst{40.9}
& \rankfirst{45.1} & \rankfirst{61.8} & \rankfirst{35.5} \\
\bottomrule
\end{tabularx}

\medskip

\begin{tabularx}{0.78\linewidth}{@{}l*{6}{>{\centering\arraybackslash}X}@{}}
\toprule
& \multicolumn{3}{c}{HuluMed-7B}
& \multicolumn{3}{c}{RadSight-8B} \\
\cmidrule(lr){2-4}
\cmidrule(l){5-7}
Setting & F1 & P & R & F1 & P & R \\
\midrule
Diagnostic VQA
& 47.6 & 32.5 & 88.6
& 52.0 & 44.4 & 62.9 \\
\midrule
Direct
& 27.7 & 58.2 & 18.2
& 42.0 & 49.6 & 36.3 \\
Report Adapt.
& 48.5 & 59.1 & 41.2
& 43.8 & \rankfirst{50.6} & 38.6 \\
Diagnostic Decomp.
& 54.9 & \ranksecond{62.0} & 49.3
& 47.7 & \ranksecond{50.5} & 45.2 \\
Diag.+Text Decomp.
& \rankfirst{57.7} & \rankfirst{62.7} & \rankfirst{53.5}
& \ranksecond{51.0} & 48.7 & \ranksecond{53.6} \\
Diag.+Text+Visual
& \ranksecond{56.6} & 60.6 & \ranksecond{53.0}
& \rankfirst{52.2} & 49.1 & \rankfirst{55.7} \\
\bottomrule
\end{tabularx}
\endgroup
\end{table}

\begin{table}[t]
\centering
\caption{Per-disease diagnostic performance across 18 disease categories on CT-RATE, evaluated using the RadBERT-based report classifier. Diseases are ordered by F1. Scores are reported as percentages.}
\label{tab:ctrate_per_disease}
\vspace{-3pt}

\begingroup
\footnotesize
\setlength{\tabcolsep}{3.8pt}
\renewcommand{\arraystretch}{1.06}

\begin{tabular}{@{}lrrr@{\hspace{10pt}}lrrr@{}}
\toprule
Disease & F1 & P & R
& Disease & F1 & P & R \\
\midrule
Pleural effusion
& 81.2 & 80.2 & 82.2
& Emphysema
& 50.6 & 45.9 & 56.4 \\

Arterial wall calcification
& 78.7 & 72.1 & 86.6
& Hiatal hernia
& 44.7 & 31.4 & 77.6 \\

Coronary artery calcification
& 74.7 & 67.3 & 83.9
& Mosaic attenuation
& 41.2 & 45.8 & 37.4 \\

Lung opacity
& 70.8 & 78.4 & 64.6
& Pulmonary fibrotic sequela
& 38.5 & 51.4 & 30.8 \\

Consolidation
& 66.3 & 65.1 & 67.5
& Lymphadenopathy
& 37.3 & 53.4 & 28.7 \\

Cardiomegaly
& 63.1 & 55.6 & 72.8
& Peribronchial thickening
& 34.5 & 45.3 & 27.9 \\

Lung nodule
& 62.8 & 56.3 & 71.1
& Interlobular septal thickening
& 33.6 & 42.8 & 27.6 \\

Pericardial effusion
& 53.6 & 49.0 & 59.0
& Bronchiectasis
& 29.4 & 43.9 & 22.1 \\

Atelectasis
& 51.5 & 42.8 & 64.6
& Medical material
& 25.2 & 73.4 & 15.2 \\
\midrule
\textbf{Micro average}
& \textbf{57.6} & \textbf{56.4} & \textbf{58.7}
& \textbf{Macro average}
& \textbf{52.1} & \textbf{55.6} & \textbf{54.2} \\
\bottomrule
\end{tabular}

\endgroup
\end{table}

The main paper reports macro-averaged metrics to give equal weight to each disease category despite class imbalance. As a complementary view, Table~\ref{tab:ct_rate_micro} reports micro-averaged F1, precision, and recall for all evaluated settings on CT-RATE.

\paragraph{Per-disease diagnostic performance.}
Table~\ref{tab:ctrate_per_disease} reports the performance of full DeCAD across the 18 CT-RATE disease categories. Macro-F1 is 52.1\%, with substantial variation across diseases. Pleural effusion achieves the highest F1 (81.2\%), followed by arterial wall calcification (78.7\%) and coronary artery wall calcification (74.7\%). Precision and recall also vary: hiatal hernia has relatively high recall (77.6\%) but low precision (31.4\%), whereas medical material has high precision (73.4\%) but low recall (15.2\%). These results highlight category-specific differences that are not captured by the aggregate score.

\subsection{\textcolor{black}{Comparison of Diagnostic Guidance Strategies}}
\label{sec:diagnostic_guidance}

\subsubsection{External-classifier guidance}

To examine alternative ways of incorporating diagnostic information into report generation, we distinguish two diagnostic-guidance settings. In \textbf{external-classifier guidance}, a separate classifier predicts diagnostic labels from the CT input, and the predicted labels are provided to the report-generation model as text prompts. In \textbf{autoregressive self-diagnosis guidance}, DeCAD first generates a diagnostic assessment from the CT input and then produces the report conditioned on both the CT input and the generated assessment.

We compare autoregressive diagnostic assessment with external-classifier guidance on CT-RATE. For the classifier-guided DeCAD variant, we train a linear classifier by linear probing on DeCAD's frozen visual encoder. During report-generator training, reference diagnostic labels are formatted as Dia-LLaMA-style prompts; at inference, the prompts are constructed from the classifier's predicted labels. The linear-probed classifier achieves an F1, precision, and recall of 53.1\%, 43.5\%, and 74.9\%, respectively. Both DeCAD variants use the same visual encoder, LLM initialization, CT-report training data, and source diagnostic labels. As shown in Table~\ref{tab:diagnostic_guidance}, autoregressive assessment achieves a macro-F1 of 52.1\%, compared with 50.7\% for classifier-guided DeCAD. It also achieves higher precision (55.6\% compared with 44.8\%), while classifier guidance achieves higher recall (67.9\% compared with 54.2\%). These results show different precision--recall trade-offs for the two guidance strategies in this setting.

These results do not establish that autoregressive self-diagnosis universally outperforms external-classifier guidance. Report-generation scores alone also do not determine how much either model relies on the CT input or its diagnostic guidance.
Prior work has explored diagnostic guidance and regional representations for CT report generation, including CT-AGRG~\cite{di2025ct}, Dia-LLaMA~\cite{chen2025dia}, and Reg2RG~\cite{chen2025large}. DeCAD complements these approaches by examining the diagnosis-to-report gap in existing MedVLMs and systematically evaluating diagnostic, regional textual, and regional visual decomposition. The resulting framework uses standard supervised fine-tuning and generates its own diagnostic assessment at inference, without requiring external diagnostic labels. Its benefits also extend to a standard VLM without medical VQA pretraining.

\definecolor{rankpurple}{HTML}{A486C8}
\definecolor{ranksecondblue}{HTML}{7FA8D4}

\begin{table*}[t]
\centering
\color{black}
\caption{
\blue{\textbf{Comparison of diagnostic guidance strategies on CT-RATE.}
F1/P/R are macro-averaged diagnostic metrics. MET, RaTE, RG, and GR
denote METEOR, RaTEScore, RadGraph-F1, and GREEN. All scores are
reported as percentages; bold marks the best result in each column.}}
\label{tab:diagnostic_guidance}
\vspace{-3pt}

\begingroup
\footnotesize
\setlength{\tabcolsep}{4.5pt}
\renewcommand{\arraystretch}{1.12}

\begin{tabular}{@{}ll ccc cc ccc@{}}
\toprule
& & \multicolumn{3}{c}{Diagnostic}
& \multicolumn{2}{c}{Text}
& \multicolumn{3}{c}{Clinical} \\
\cmidrule(lr){3-5}
\cmidrule(lr){6-7}
\cmidrule(l){8-10}
Method & Guidance
& F1 & P & R
& MET & BLEU
& RaTE & RG & GR \\
\midrule

Dia-LLaMA
& Classifier
& 30.4 & 45.8 & 26.2
& 24.0 & 24.3
& 66.2 & \textbf{25.7} & 41.7 \\

DeCAD
& Classifier
& 50.7 & 44.8 & \textbf{67.9}
& 21.6 & 19.7
& 65.8 & 22.4 & 37.6 \\

DeCAD
& Autoregressive
& \textbf{52.1} & \textbf{55.6} & 54.2
& \textbf{26.3} & \textbf{27.1}
& \textbf{66.5} & 23.6 & \textbf{42.0} \\

\bottomrule
\end{tabular}
\endgroup
\vspace{-3pt}
\end{table*}

\begin{table}[ht]
\centering
\color{black}
\caption{\blue{Disease-list prompting and diagnostic decomposition on CT-RATE (\%). Within each model and metric, the best score is shaded purple and bold; the second-best distinct score is shaded blue and underlined.}}
\label{tab:disease_list_prompting}

\begingroup
\footnotesize
\setlength{\tabcolsep}{3pt}
\renewcommand{\arraystretch}{1.10}
\def\rankfirst#1{\cellcolor{rankpurple!30}\textbf{#1}}
\def\ranksecond#1{\cellcolor{ranksecondblue!30}\underline{#1}}

\begin{tabularx}{\linewidth}{@{}lc*{9}{>{\centering\arraybackslash}X}@{}}
\toprule
& & \multicolumn{3}{c}{CT-CHAT}
& \multicolumn{3}{c}{MedGemma-1.5-4B}
& \multicolumn{3}{c}{Lingshu-7B} \\
\cmidrule(lr){3-5}
\cmidrule(lr){6-8}
\cmidrule(l){9-11}
Setting & Fine-tuning
& F1 & P & R
& F1 & P & R
& F1 & P & R \\
\midrule

Direct & --
& 17.8 & 37.4 & 18.1
& 13.7 & 39.4 & 12.4
& 7.1 & 28.6 & 5.4 \\

Disease Prompt & --
& \ranksecond{27.2} & 24.4 & \rankfirst{40.7}
& \ranksecond{20.6} & 23.7 & \rankfirst{38.2}
& 10.8 & 15.0 & 17.6 \\

Report Adapt. & $\checkmark$
& 17.7 & \ranksecond{39.7} & 15.1
& 19.8 & \ranksecond{46.0} & 19.5
& \ranksecond{21.5} & \rankfirst{56.2} & \ranksecond{17.8} \\

Diagnostic Decomp. & $\checkmark$
& \rankfirst{30.7} & \rankfirst{43.9} & \ranksecond{30.1}
& \rankfirst{32.3} & \rankfirst{50.7} & \ranksecond{28.9}
& \rankfirst{30.8} & \ranksecond{47.2} & \rankfirst{27.8} \\

\bottomrule
\end{tabularx}

\medskip

\begin{tabularx}{0.78\linewidth}{@{}lc*{6}{>{\centering\arraybackslash}X}@{}}
\toprule
& & \multicolumn{3}{c}{HuluMed-7B}
& \multicolumn{3}{c}{RadSight-8B} \\
\cmidrule(lr){3-5}
\cmidrule(l){6-8}
Setting & Fine-tuning & F1 & P & R & F1 & P & R \\
\midrule

Direct & --
& 21.3 & 52.5 & 15.4
& 34.6 & \ranksecond{43.9} & 30.9 \\

Disease Prompt & --
& 36.1 & 31.0 & \rankfirst{59.5}
& \ranksecond{38.0} & 37.1 & \rankfirst{40.7} \\

Report Adapt. & $\checkmark$
& \ranksecond{37.7} & \ranksecond{53.8} & 33.2
& 37.2 & 43.8 & 33.7 \\

Diagnostic Decomp. & $\checkmark$
& \rankfirst{45.0} & \rankfirst{57.4} & \ranksecond{42.9}
& \rankfirst{42.2} & \rankfirst{45.3} & \ranksecond{40.6} \\

\bottomrule
\end{tabularx}
\endgroup
\end{table}

\begin{table}[t]
\centering
\color{black}
\caption{\blue{\textbf{Diagnostic VQA with original and shuffled CT inputs on CT-RATE.} AUC is the macro-average of per-disease AUROC values computed from yes/no logits. F1/P/R are macro-averaged diagnostic metrics. All values are percentages.}}
\label{tab:vqa_shuffle}
\vspace{-3pt}

\begingroup
\footnotesize
\setlength{\tabcolsep}{5pt}
\renewcommand{\arraystretch}{1.12}

\begin{tabular}{@{}lcccc cccc@{}}
\toprule
& \multicolumn{4}{c}{Original image}
& \multicolumn{4}{c}{Shuffled image} \\
\cmidrule(lr){2-5}
\cmidrule(l){6-9}
Model & AUC & F1 & P & R & AUC & F1 & P & R \\
\midrule
CT-CHAT & 69.7 & 31.4 & 22.7 & 85.0 & 51.4 & 26.7 & 17.8 & 80.0 \\
MedGemma-1.5 & 70.1 & 20.6 & 33.7 & 27.9 & 53.2 & 12.1 & 16.2 & 18.5 \\
Lingshu-7B & 64.5 & 24.6 & 27.5 & 59.3 & 51.4 & 20.8 & 19.3 & 50.6 \\
HuluMed-7B & 80.9 & 46.3 & 33.4 & 87.2 & 51.9 & 27.0 & 19.3 & 52.5 \\
RadSight-8B & 82.3 & 49.7 & 45.1 & 66.8 & 51.8 & 20.7 & 19.3 & 27.3 \\
\bottomrule
\end{tabular}
\endgroup
\vspace{-3pt}
\end{table}

\begin{table}[t]
\centering
\color{black}
\caption{\blue{\textbf{Always-positive VQA baseline on CT-RATE.} Scores are macro-F1 percentages.}}
\label{tab:vqa_always_positive}
\vspace{-3pt}

\begingroup
\footnotesize
\setlength{\tabcolsep}{5pt}
\renewcommand{\arraystretch}{1.12}

\begin{tabular}{@{}lcccc@{}}
\toprule
Model & VQA & Direct Report & VQA gap & Always-positive gap \\
\midrule
CT-CHAT & 31.4 & 17.8 & 13.6 & 13.2 \\
MedGemma-1.5 & 20.6 & 13.7 & 6.9 & 17.3 \\
Lingshu-7B & 24.6 & 7.1 & 17.5 & 23.9 \\
HuluMed-7B & 46.3 & 21.3 & 25.0 & 9.7 \\
RadSight-8B & 49.7 & 34.6 & 15.1 & $-3.6$ \\
\bottomrule
\end{tabular}
\endgroup
\vspace{-3pt}
\end{table}

\begin{table}[t]
\centering
\caption{Effect of diagnostic supervision coverage on whole-image report generation. P, R, and F1 are micro- or macro-averaged diagnostic metrics, reported as percentages. Disease-specific Diagnose supervision is applied to randomly selected subsets of the training cases.}
\label{tab:diagnostic_granularity}
\small
\setlength{\tabcolsep}{5pt}
\begin{tabular}{lrrrrrr}
\toprule
& \multicolumn{3}{c}{Micro} & \multicolumn{3}{c}{Macro} \\
\cmidrule(lr){2-4}\cmidrule(lr){5-7}
Setting & P & R & F1 & P & R & F1 \\
\midrule
Whole-image baseline
  & 44.9 & 22.2 & 29.7 & 37.6 & 15.1 & 15.9 \\
Normal/abnormal assessment
  & 44.8 & 22.1 & 29.6 & 24.4 & 13.9 & 13.9 \\
Diagnose on 1/3 of cases
  & 53.7 & 31.3 & 39.6 & 43.6 & 22.3 & 25.1 \\
Diagnose on 2/3 of cases
  & 53.7 & 37.5 & 44.1 & 44.0 & 28.6 & 30.2 \\
Diagnose on all cases
  & 52.4 & 47.3 & \textbf{49.7}
  & 45.7 & 39.0 & \textbf{39.0} \\
\bottomrule
\end{tabular}
\end{table}

\subsubsection{Disease list prompting guidance}

We added a disease-specific prompting control that provides the same disease checklist as diagnostic VQA while keeping model weights and CT inputs fixed. As shown in Table~\ref{tab:disease_list_prompting}, disease-list prompting improves macro-F1 for all five models, including from 17.8\% to 27.2\% for CT-CHAT and from 21.3\% to 36.1\% for HuluMed-7B. Without additional fine-tuning, it outperforms Report Adaptation for three of the five models. Diagnostic Decomposition nevertheless achieves higher macro-F1 than disease-list prompting for every model, including 30.7\% for CT-CHAT and 45.0\% for HuluMed-7B. Therefore, disease-specific cues can contribute to the diagnosis-to-report gap, but explicit diagnostic supervision further improves the model’s ability to generate diagnostically complete reports beyond disease-specific prompting alone.

\subsection{\textcolor{black}{Additional Analysis of Diagnostic VQA}}
\label{sec:vqa_analysis}

\subsubsection{Whether VQA captures vision-conditioned disease discrimination?}

To assess whether VQA reflects vision-conditioned disease discrimination rather than merely the effect of explicit disease-specific prompting, we evaluate both the model’s continuous confidence and its final yes/no decisions under original and shuffled CT inputs. In disease-specific VQA, the model is given a CT volume and a disease query and determines whether the queried disease is present.
First, to obtain a continuous VQA confidence score, we apply a softmax to the logits for the constrained “yes” and “no” responses and use the resulting “yes” probability. We then compute AUROC for each disease category and macro-average across categories. AUROC measures confidence-based discriminability, i.e., whether positive cases receive higher “yes” confidence than negative cases.  In contrast, macro-F1 measures operational binary diagnostic quality, i.e., whether the model converts its evidence into correct final yes/no decisions, including the resulting precision-recall trade-off. As shown in Table~\ref{tab:vqa_shuffle}, the original-image macro-AUCs are 69.7\%, 70.1\%, 64.5\%, 80.9\%, and 82.3\% for CT-CHAT, MedGemma, Lingshu, HuluMed, and RadSight, respectively.

Second, we add a shuffled-image control (Table~\ref{tab:vqa_shuffle}). We retain each disease query, ground-truth label, and evaluation protocol, while randomly replacing the paired CT volume with that of another patient, thereby breaking image-label correspondence. We report both the original- and shuffled-image macro-AUC, together with F1, precision, and recall. After shuffling, macro-AUC decreases by 18.3\%, 16.9\%, 13.1\%, 29.0\%, and 30.5\% for CT-CHAT, MedGemma, Lingshu, HuluMed, and RadSight, respectively. F1 also decreases for all models, with the largest drops for HuluMed (19.3\%) and RadSight (29.0\%). Thus, both the continuous confidence scores and final yes/no decisions depend on the visual input.

Together, these analyses show that VQA captures vision-conditioned disease discrimination. Accordingly, VQA macro-F1 evaluates models’ end-to-end vision-conditioned diagnostic decisions, jointly reflecting their discrimination of positive versus negative cases and their conversion of this evidence into final yes/no outputs.

\subsubsection{How the disease-specific yes/no formulation affects the observed VQA-report difference?}

We add an always-positive VQA baseline (Table~\ref{tab:vqa_always_positive}) to assess whether an affirmative response tendency alone can reproduce the observed VQA-report difference. Although this image-free strategy attains a non-trivial macro-F1 of 31.0\%, it does not consistently reproduce the observed gaps across models. For CT-CHAT, the always-positive and observed gaps are similar (13.2\% vs. 13.6\%), consistent with its high-recall/low-precision VQA behavior. This tendency is itself part of the model’s VQA diagnostic behavior: it reflects how the model converts visual evidence into a final yes/no decision, with high sensitivity but limited specificity indicating less selective disease discrimination. However, for HuluMed and RadSight, the observed gaps substantially exceed the always-positive gaps (25.0\% vs. 9.7\% and 15.1\% vs. -3.6\%, respectively). Thus, although model-specific yes/no response behavior is a meaningful component of VQA performance and can contribute to the VQA-report difference, an indiscriminate affirmative strategy cannot universally account for the observed gaps, particularly for HuluMed and RadSight.

\subsubsection{Whether different label-extraction procedures confound the comparison?}

We clarify the comparability of the label-extraction procedures. The VQA and report-based evaluations use the same CT studies, disease categories, and reference binary labels; they differ only in how model predictions are converted into disease labels. For report-based evaluation, we extract diagnostic labels from generated reports using RadBERT, following~\cite{hamamci2026generalist}. On CT-RATE, we validated RadBERT against the official diagnostic labels, achieving an F1 score of 98.1\%, precision of 99.4\%, and recall of 96.9\%. On Merlin, our independently trained ModernBERT extractor achieves a macro-AUC of 98.8\% and macro-F1 of 95.3\% on the held-out test set across 30 disease categories (Appendix Table~\ref{tab:merlin_extractor}). Thus, report-side label extraction introduces limited measurement error and yields labels closely aligned with the official labels used for VQA evaluation. The observed VQA-report differences are therefore unlikely to be primarily explained by the use of different label-extraction procedures.

Taken together, these analyses show that VQA is a visually grounded, disease-specific probe of diagnostic discrimination. The VQA-report difference should nevertheless be interpreted as task-dependent and in light of model-specific yes/no decision behavior.

\subsection{Coverage of Diagnostic Supervision}
\label{sec:diagnostic_granularity}

\paragraph{How much diagnostic supervision is needed?} The preceding results show that diagnostic decomposition improves report generation, but leave open how broadly disease-specific diagnostic supervision must be applied during training. We examine this question using the whole-image baseline and four assessment-then-report settings. The coarsest assessment states only whether the CT is normal or abnormal. For disease-specific diagnostic supervision, we randomly select approximately one-third, two-thirds, or all training cases to receive a full positive-finding assessment before report generation. At inference, the model generates its own assessment before generating the report. 

Table~\ref{tab:diagnostic_granularity} shows that a binary normal/abnormal assessment does not improve on the whole-image baseline: micro-F1 is 29.6\% versus 29.7\%, and macro-F1 is 13.9\% versus 15.9\%. Compressing diverse findings into a single judgment therefore provides little disease-specific guidance for subsequent report generation. When disease-specific Diagnose supervision is introduced for one-third of the training cases, micro-precision increases from 44.9\% to 53.7\% and macro-precision from 37.6\% to 43.6\%, indicating that explicit diagnostic supervision improves the precision of the resulting reports even when applied to only a subset of the training data. As supervision coverage increases, recall rises steadily: micro-recall increases from 31.3\% to 37.5\% and 47.3\%, while macro-recall increases from 22.3\% to 28.6\% and 39.0\% for one-third, two-thirds, and all training cases, respectively. Precision remains broadly stable across these three settings. Together, these results suggest that disease-specific diagnostic supervision primarily improves report precision when first introduced, while broader supervision coverage progressively improves recall and reduces diagnostic omissions.

\subsection{Qualitative Analysis}
\label{sec:qualitative_analysis}

\begin{figure}
    \centering
    \includegraphics[width=\linewidth,height=0.8\textheight,keepaspectratio]{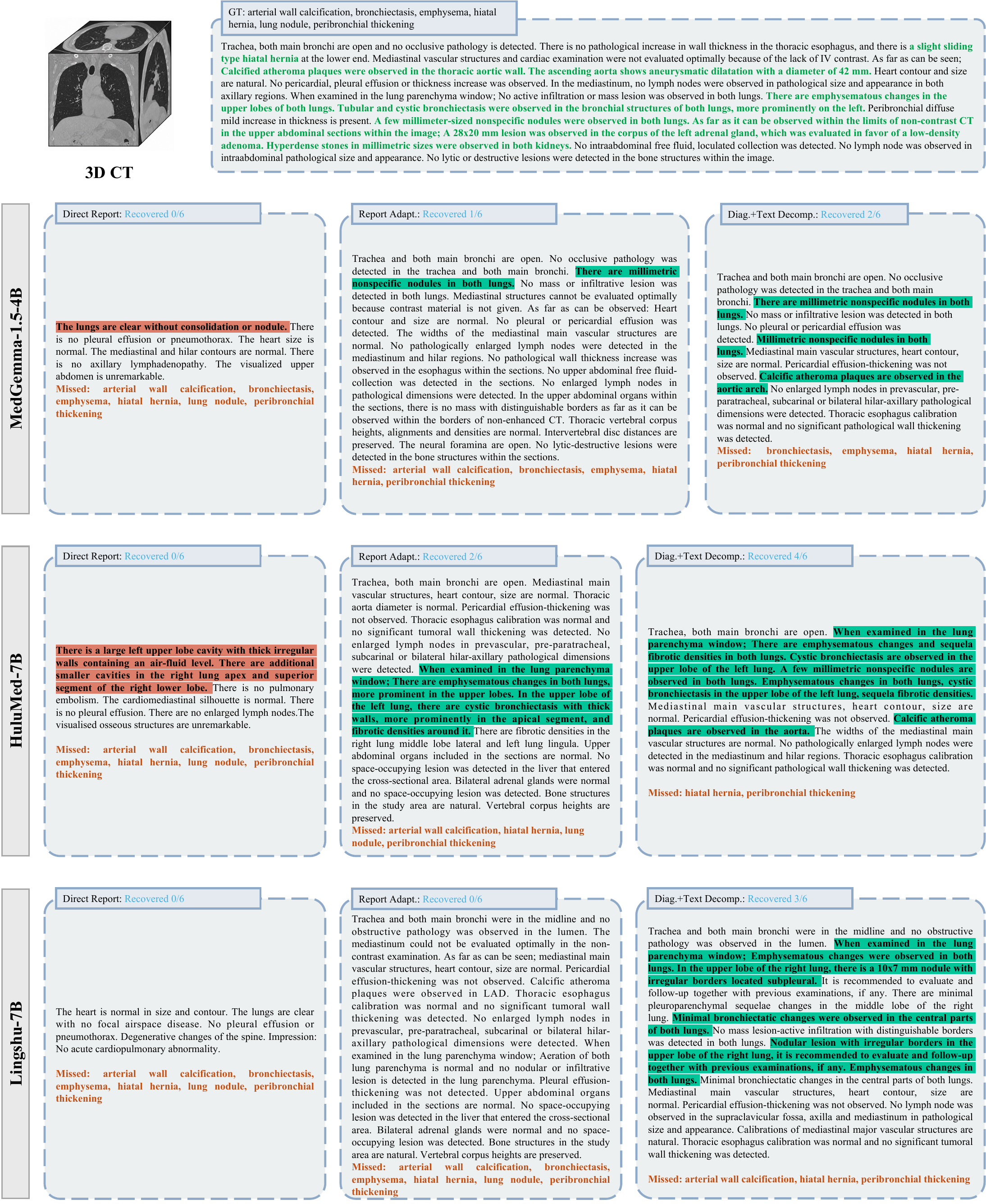}
\caption{\textbf{Qualitative comparison on CT-RATE across MedGemma-1.5-4B, HuluMed-7B, and Lingshu-7B.}
For the same CT study with six ground-truth positive findings, we compare Direct Report, Report Adaptation, and diagnostic + textual decomposition.
Green highlights mark generated descriptions corresponding to ground-truth findings, orange text lists ground-truth findings omitted from each report, and red highlights indicate additional findings not described in the reference report.
The header of each panel reports the number of ground-truth findings recovered.
Across all three models, diagnostic + textual decomposition recovers more reference findings than the corresponding direct and adapted reports.}
    \label{fig:report_case_01}
\end{figure}

\begin{figure}
    \centering
    \includegraphics[width=\linewidth,height=0.8\textheight,keepaspectratio]{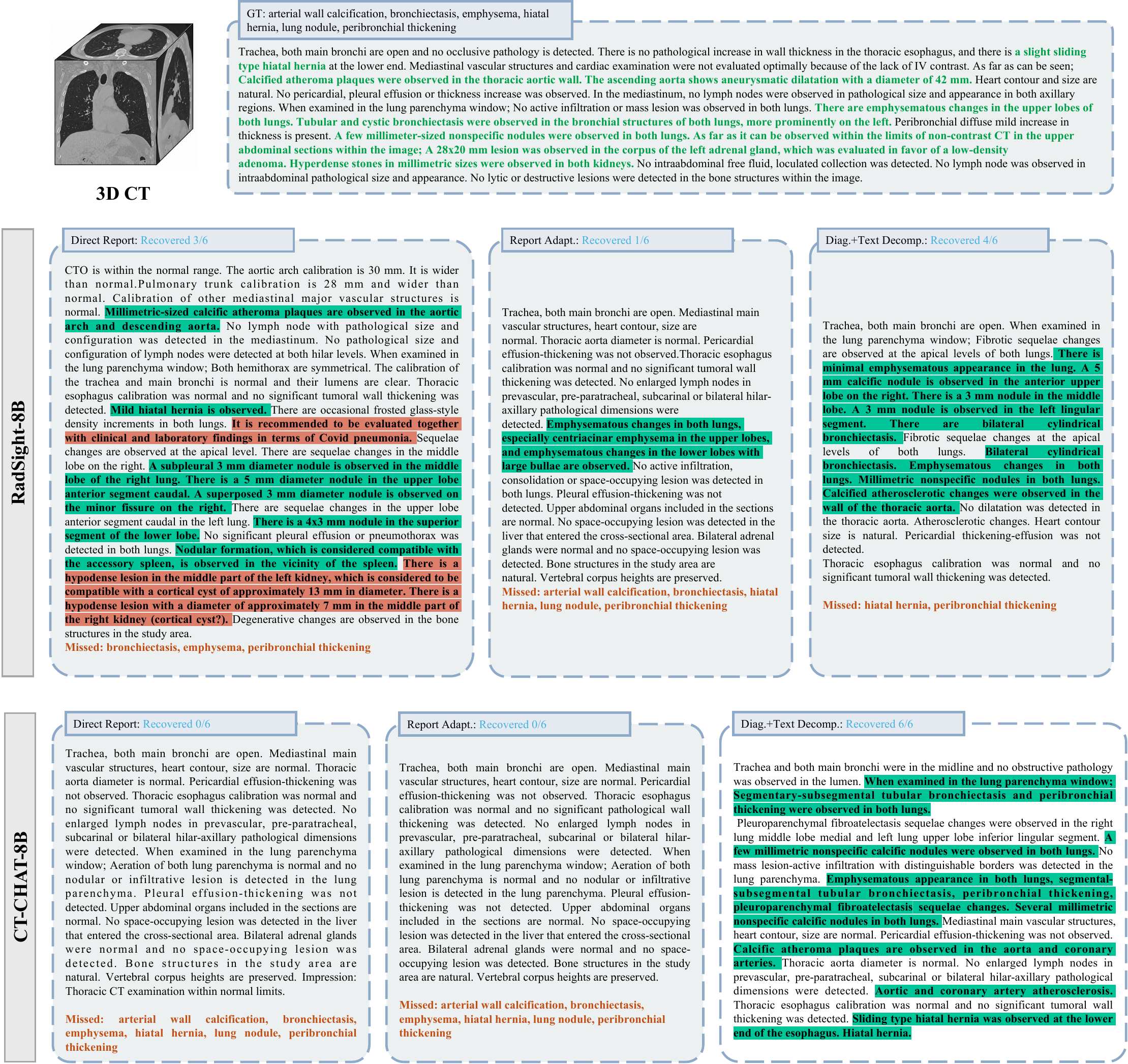}
\caption{\textbf{Qualitative comparison on CT-RATE across RadSight-8B and CT-CHAT-8B.}
We compare Direct Report, Report Adaptation, and diagnostic + textual decomposition on the same CT study with six ground-truth positive findings.
Green highlights mark generated descriptions corresponding to ground-truth findings, orange text lists ground-truth findings omitted from each report, and red highlights indicate additional findings not described in the reference report.
The header of each panel reports the number of ground-truth findings recovered.
Diagnostic + textual decomposition increases the recovery of reference findings for both models, while conventional report adaptation does not consistently reduce diagnostic omissions.}
    \label{fig:report_case_02}
\end{figure}

Figures~\ref{fig:report_case_01} and~\ref{fig:report_case_02} provide a qualitative comparison of Direct Report, Report Adaptation, and combined diagnostic and regional textual decomposition across five MedVLMs on the same CT study.
The reference report contains six positive findings: arterial wall calcification, bronchiectasis, emphysema, hiatal hernia, lung nodule, and peribronchial thickening. Direct report generation frequently omits these abnormalities despite producing fluent radiological text. Conventional Report Adaptation recovers some findings for several models, but substantial omissions remain, and the improvement is not consistent across models. In contrast, diagnostic + textual decomposition increases the number of ground-truth findings expressed in the final report for all five MedVLMs. The recovered abnormalities span multiple anatomical regions and include pulmonary nodules, emphysematous and bronchiectatic changes, arterial calcification, hiatal hernia, and peribronchial thickening. 
These examples complement the quantitative results by illustrating the diagnosis-to-report gap at the level of individual findings: diagnostically relevant abnormalities can be omitted during free-form report composition, whereas explicitly decomposing diagnosis and description helps express more diagnostically relevant findings in the generated report.

\appendix

\end{document}